\documentclass[10pt,journal,compsoc]{IEEEtran}
\usepackage{epsfig}
\usepackage{graphicx}
\usepackage{subfigure}
\graphicspath{{./fig}}
 \DeclareGraphicsExtensions{.eps,.pdf,.jpg,.png}
\usepackage{amsmath}
\usepackage{amssymb}
\usepackage{multirow}
\usepackage{tabularx}
\usepackage{diagbox}
\usepackage{bm}
\usepackage{algorithm}
\usepackage{algorithmic}
\usepackage[breaklinks=true,bookmarks=false]{hyperref}

\def\E{{\rm E}}

\def\N{{\rm N}}
\def\T{{\top}}

\ifCLASSOPTIONcompsoc
  \usepackage[nocompress]{cite}
\else
  \usepackage{cite}
\fi
\ifCLASSINFOpdf
\else
\fi
\begin{document}
%
\title{Deformable Generator Networks: Unsupervised Disentanglement of Appearance and Geometry}
%
%
%
%

\author{Xianglei~Xing,~\IEEEmembership{Member,~IEEE,}
        Ruiqi~Gao,
        Tian~Han,
        Song-Chun~Zhu,~\IEEEmembership{Fellow,~IEEE,}
        and Ying Nian~Wu
\IEEEcompsocitemizethanks{\IEEEcompsocthanksitem X. Xing is with the College of Intelligent System Science and Engineering, Harbin Engineering University, Heilongjiang,
China, 150001.\protect

E-mail: xingxl@hrbeu.edu.cn

\IEEEcompsocthanksitem T. Han is with the Computer Science Department, Stevens Institute of Technology, Hoboken, New Jersey, 07030.\protect

E-mail: than6@stevens.edu

\IEEEcompsocthanksitem R. Gao, S.-C. Zhu and Y. N. Wu are with Department of statistics, University of California, Los Angeles,  California 90095.\protect\\
E-mail: \{ruiqigao,sczhu,ywu\}@stat.ucla.edu}
\thanks{IEEE Transactions on Pattern Analysis and Machine Intelligence, VOL. 44, NO. 3, MARCH 2022}}

\IEEEtitleabstractindextext{%
\begin{abstract}
We present a deformable generator model to disentangle the appearance and geometric information for both image and video data in a purely unsupervised manner. The appearance generator network models the information related to appearance, including color, illumination, identity or category,  while the geometric generator performs geometric warping, such as rotation and stretching, through generating deformation field which is used to warp the generated appearance to obtain the final image or video sequences. Two generators take independent latent vectors as input to disentangle the appearance and geometric information from image or video sequences. For video data, a nonlinear transition model is introduced to both the appearance and geometric generators to capture the dynamics over time. The proposed scheme is general and can be easily integrated into different generative models. An extensive set of qualitative and quantitative experiments shows that the appearance and geometric information can be well disentangled, and the learned geometric generator can be conveniently transferred to other image datasets that share similar structure regularity to facilitate knowledge transfer tasks.
\end{abstract}

}

\maketitle

\IEEEdisplaynontitleabstractindextext

%
\IEEEpeerreviewmaketitle

\IEEEraisesectionheading{\section{Introduction}\label{sec:introduction}}

%
%
%
%
\IEEEPARstart{L}{earning} disentangled structures of the observations \cite{bengio2013representation,mathieu2016disentangling} is a fundamental problem towards controlling modern deep models and understanding the world. Conceptual understanding requires a disentangled representation that separates the underlying explanatory factors and shows the important attributes of the real-world data explicitly \cite{burgess2018understanding,achille2017emergence}. For instance, given an image dataset of human faces, a disentangled representation can separate the face's appearance attributes, such as color, light source, identity, gender, and the geometric attributes, such as face shape and viewing angle. Such disentangled representations are semantically meaningful not only in building more transparent and interpretable generative models, but also useful for a large variety of downstream AI tasks such as transfer learning and zero-shot inference where humans excel but machines struggle \cite{lake2017building}. It has also been shown that such disentangled representations are more generalizable and robust against adversarial attacks \cite{alemi2016deep}.

Recently, deep generative models, e.g., generator model, have shown great promise in learning representation of images \cite{brock2018large,karras2019style}.  Most efforts focus on developing sophisticated architectures and training paradigms for sharp and realistic-looking image synthesis~\cite{lucic2017gans,BigGAN,karras2019style,han2019divergence}. However, the learned latent representation is often entangled and not interpretable. Learning disentangled and interpretable representation without supervision for deep generative models is challenging, e.g., from face images where no facial landmarks are given. We are particularly interested in learning the disentangled representation for the generative models, since it allows users to specify the desired properties of the output by controlling the generative factors encoded in each latent dimension. Many exciting applications require generative models that can synthesize novel instances while certain key factors are held fixed, for example, generating a face image with desired attributes, such as color, face shape, expression and view (which can be learned and transferred from another person), while keeping the identity fixed. There are increasing demands on such generative models in various domains, such as image manipulation \cite{lee2018diverse, huang2018multimodal}, video generation~\cite{xie2019motion}, multi-view learning~\cite{han2018learning}, and machine learning fairness  \cite{NIPS2019_9603}.

In this paper, we propose a deformable generator model that disentangles the appearance and geometric information and is learned in a purely unsupervised manner under a unified probabilistic framework. Specifically, our model integrates two generator networks: one appearance generator and one geometric generator with  two sets of independent latent factors. The dense deformation fields (displacements of the coordinates of each pixel) are generated by the geometric generator, which act on the image intensities generated by the appearance generator to obtain the final image through a differentiable warping function. The model is learned by alternating back-propagation through the model parameters and latent variables of the two networks. The proposed model can be applied to both image and video data. For the spatial-temporal process in video data, the proposed dynamic deformable model introduces non-linear transition models for the latent vectors. The transition model for the appearance factors can generate dynamic textures which represent non-trackable motion, and the transition model for the geometric factors can generate intuitive physics which represent trackable motion. The proposed method can learn well-disentangled representation. The learned geometric representations can be transferred to other datasets that share similar structure regularity and can be used for various tasks.

Our contributions are summarized below:
\begin{itemize}
\item We propose a deformable generator network to disentangle the appearance and geometric information in a purely unsupervised manner.
\item We propose a dynamic deformable generator network to disentangle the appearance and geometric information for the spatial-temporal process in video data.
\item We perform extensive experiments both qualitatively and quantitatively which show that the appearance and geometric information can be well disentangled and effectively transferred to other datasets and tasks.
\end{itemize}

\section{Related work}

Existing work on learning disentangled representation using deep generative models generally fall into two categories: implicit learning and explicit learning.

The implicit learning methods encourage the disentanglement of latent factors by adding regularization terms in the loss function, which fall into two categories: the Generative Adversarial Networks (GANs) \cite{goodfellow2014generative,li2018unsupervised,tran2017disentangled} and the Variational Auto-encoders (VAEs) \cite{kingma2013auto,RezendeICML2014,kumar2017variational}. InfoGAN \cite{chen2016infogan} and $\beta$-VAE \cite{higgins2016beta} are representatives for the two families respectively, which focus on designing loss functions to encourage the independence of latent factors. More specifically, InfoGAN \cite{chen2016infogan}, which belongs to the GANs family, is proposed under the principle of maximization of the mutual information between the observations and a subset of latent factors.  However, its disentangling performance is sensitive to the choice of the prior and the number of latent factors. 
$\beta$-VAE \cite{higgins2016beta}, from the VAEs family, learns disentangled representations by utilizing a VAE objective but with a stronger penalty on the discrepancy between the posterior distributions of the latent factors and independent Gaussian priors, making latent factors to be independent as much as possible, thus giving a more robust and stable solution for disentanglement. 
Recently, StyleGAN \cite{karras2019style}, StyleGAN2 \cite{karras2020analyzing}, ProGAN \cite{karras2017progressive} are proposed for generating high resolution images. StyleGAN2 can separate fine-grained variation (e.g., hair, freckles) from high-level features (e.g., pose, identity) by mapping the latent code to layer-wise style codes and then fed into each convolutional layer with Adaptive Instance Normalization (AdaIN), but it does not provide explicit control over these elements. HoloGAN \cite{nguyen2019hologan} employs the basic structure of StyleGAN and further utilizes rigid-body transformation of the latent feature space to disentangle the pose and the identity information of the generated objects. However, it can not disentangle the shape and identity information well. Although these implicit methods can be learned unsupervisedly, the learned representation is not controllable and not well disentangled.

The explicit methods, on the other hand, model appearance and geometric explicitly by separate models, originated from the Active Appearance Models (AAM) which \cite{hallinan1999two,cootes2001active,kossaifi2017fast} learn the appearance and geometric information by performing principal component analysis (PCA) on appearance and facial landmarks separately. Unlike the AAM method which requires hand-annotated facial landmarks, our proposed deformable generator model is purely unsupervised and learns from images or videos alone. Recently, \cite{kossaifi2017gagan} incorporates the shape geometry into the GANs which generalizes the linear AAM model to the nonlinear model to learn well separated appearance and geometric information. However, this method \cite{kossaifi2017gagan}  also requires annotated facial landmarks for each image during training. Unsupervised disentanglement of the appearance and geometric information is challenging and remains not well-explored. \cite{shu2018deforming} follows this direction, but their model focused on the auto-encoder (AE) only, and cannot generate new images with desired attributes by controlling the latent factors. Moreover, it is not developed under probabilistic framework as ours.



\section{Model and learning algorithm}
This section provides the details of the model for 2D-image data and the corresponding learning and inference algorithm. The dynamic deformable model for 3D-video data will be introduced in the next section.
\subsection{Model}
\begin{figure}[h]
\begin{center}
\centerline{\includegraphics[width=\columnwidth]{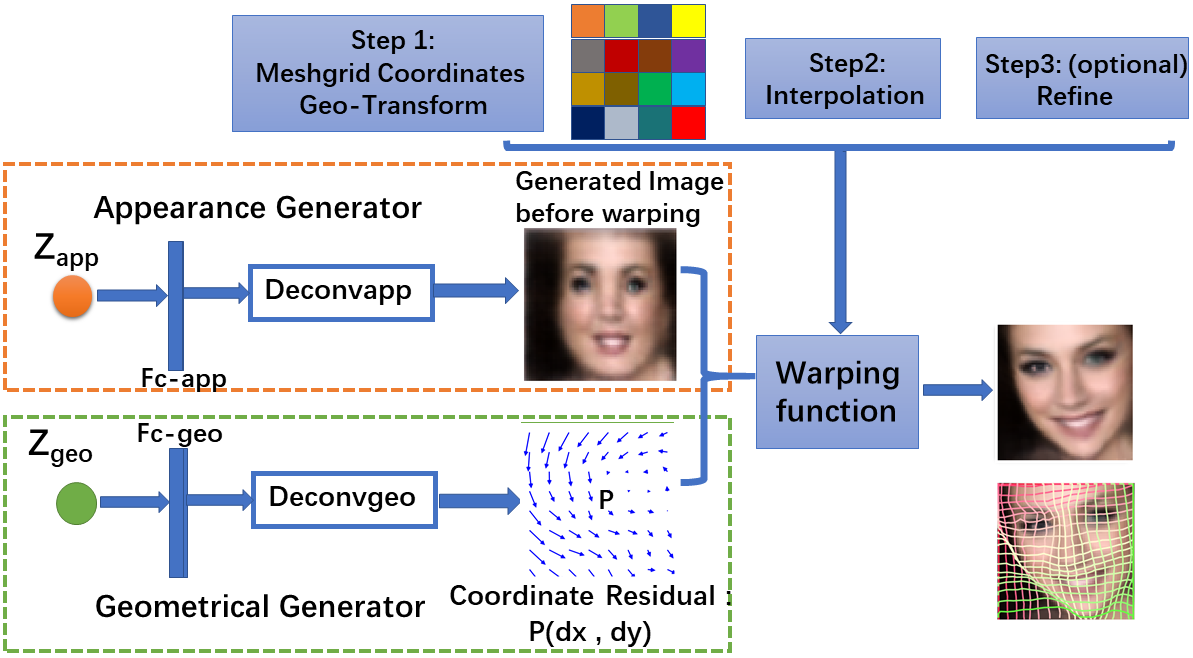}}
\caption{An illustration of the proposed model. The model contains two generator networks: one appearance generator and one geometric generator. The two generators are combined by a warping function to produce the final image. The warping function includes a geometric transformation operation for image coordinates and a differentiable interpolation operation. The refining operation is optional for improving the performance of the warping function.}
\label{fig:basicmodel}
\end{center}
\end{figure}
The proposed model contains two generator networks: one appearance generator and one geometric generator, which are combined by a warping function to produce the final images or video frames, as shown in figure \ref{fig:basicmodel}. Suppose an arbitrary image or video frame $X \in \mathbb{R}^{D_x \times D_y \times 3}$  is generated with two independent latent vectors, $Z^a \in \mathbb{R}^{d_a}$ which controls the appearance, and $Z^g \in \mathbb{R}^{d_g}$ which controls the geometric information. Varying the geometric latent vector $Z^g$ and fixing the appearance latent vector $Z^a$, we can transform an object's geometric information, such as rotating it with certain angle and changing its shape. On the other hand, varying $Z^a$ and fixing $Z^g$, we can change the identity or the category of the object, while keeping it geometric information unchanged, such as the same viewing angle or the same shape. 

The model can be expressed as
\begin{eqnarray}
X &=&F(Z^a,Z^g; \theta) + \epsilon \nonumber\\
&=& F_w(F_a(Z^a;\theta_a),F_g(Z^g;\theta_g)) + \epsilon
\label{eq:wholemodel}
\end{eqnarray}
where $Z^a \sim {\rm N}(0, I_{d_a})$, $Z^g \sim {\rm N}(0, I_{d_g})$, and $\epsilon \sim {\rm N}(0, \sigma^2 I_D)$ ($D = D_x \times D_y \times 3$) are independent.
$F_w$ is the warping function, which uses the deformation field generated by the geometric generator $F_g(Z^g;\theta_g)$ to warp the image generated by the appearance generator $F_a(Z^a;\theta_a)$ to synthesize the final output image $X$.

\subsection{Warping function}
A warping function usually includes a geometric transformation operation for image coordinates and a differentiable interpolation (or resampling) operation. The geometric transformation describes the target coordinate $(x,y)$ for every location $(u,v)$ in the source coordinate. The geometric operation only modifies the positions of pixels in an image without changing the color or illumination. Therefore, the appearance information and the geometric information are naturally disentangled by the two generators in the proposed model.

The geometric transformation $\Phi$ can be a rigid affine mapping, as used in the spatial transformer networks \cite{jaderberg2015spatial}, or a non-rigid deformable mapping, which is the case in our work. Specifically, the coordinate displacement $(dx,dy)$ (or the dense optical flow field) of each regular grid $(x,y)$ in the output warped image $X$ are generated by the geometric generator $F_g(Z^g;\theta_g)$. The  point-wise transformation in this deformable mapping can be formulated as
\begin{eqnarray}
{u \choose v} =  \Phi_{(Z^g,\theta_g)}{x \choose y} = {x + dx \choose y + dy}
\label{eq:geotransform}
\end{eqnarray}
where $(u,v)$ are the source coordinates of the image generated by the appearance generator $F_a(Z^a;\theta_a)$.

Since the evaluated $(u,v)$ by Eq.(\ref{eq:geotransform}) do not always have integer coordinates, each pixel's value of the output warped image $X$  can be computed by a differentiable interpolation operation. Let $X_a=F_a(Z^a;\theta_a)$ denote the image generated by the appearance generator.  The warping function $F_w$ can be formulated as
 \begin{equation}
X(x,y) = F_I( X_a(x+dx,y+dy)),
\end{equation}
where $F_I$ is the differentiable interpolation function. We use a differentiable bilinear interpolation:
\begin{equation}
X(x,y) = \sum_j^{D_y}\sum_i^{D_x} X_a(i,j)M(1-|u-i|)M(1-|v-j|)
\end{equation}
where $M(\cdot)=\max(0,\cdot)$.  The details of back-propagation through this bilinear interpolation can be found in \cite{jaderberg2015spatial}.

The displacement $(dx, dy)$  is used in the deformable convolutional networks \cite{dai2017deformable}. The computation of coordinates displacement $(dx,dy)$ is  known as the optical flow estimation \cite{sun2014quantitative,ilg2017flownet}. Our work is concerned with modeling and generating the optical flow, in addition to estimating the optical flow.

Notice that the displacement $(dx, dy)$ may also indicates the motion of the objects in the scene, or the change of viewpoint relative to 3D objects in the scene. It is natural to incorporate motion and 3D models into the geometric generator where the change or variation of $Z^g$ depends on the motion and 3D information.

\subsection{Inference and learning}
To learn this deformable generator model, we introduce a learning and inference algorithm for two latent vectors, without designing and learning extra inference networks. Our method is motivated by a maximum likelihood learning algorithm for generator networks \cite{han2017alternating}. Specifically, the proposed model can be trained by maximizing the log-likelihood on the training dataset $\{X_i,i = 1,\dots,N\}$,
\begin{eqnarray}
L(\theta) &=& \frac{1}{N}\sum_{i=1}^N\log p(X_i; \theta)  \nonumber \\
&=& \frac{1}{N} \sum_{i=1}^N \log \int p(X_i, Z^a_i, Z^g_i; \theta)dZ^a_idZ^g_i,
\label{eq:MLE}
\end{eqnarray}
where we integrate out the uncertainties of $Z^a_i$  and $Z^g_i$ in the complete-data log-likelihood to get the observed-data log-likelihood.

We can evaluate the gradient of $L(\theta)$ by the following well-known result, which is related to the EM algorithm:
\begin{eqnarray}
&&\frac{\partial}{\partial \theta} \log p(X; \theta)\nonumber\\
&&=  \frac{1}{p(X; \theta)}\frac{\partial}{\partial \theta} \int p(X, Z^a, Z^g)dZ^adZ^g \nonumber \\
&&=\E_{p(Z^a,Z^g|X; \theta)} \left[\frac{\partial}{\partial \theta} \log p(X, Z^a,Z^g; \theta) \right]
\label{eq:expectposterior}
\end{eqnarray}
Since the expectation in Eq.(\ref{eq:expectposterior}) is usually analytically intractable, we employ Langevin dynamics to draw samples from the posterior distribution $p(Z_a,Z_g|X;\theta)$ and compute the Monte Carlo average to estimate the expectation term. For each observation $X$, the latent vectors $Z^a$ and $Z^g$ can be sampled from $p(Z^a,Z^g|X;\theta)$  alternately by Langevin dynamics: we fix $Z^g$ and sample $Z^a$ from $p(Z^a|X;Z^g,\theta)$ $\propto $ $p(X, Z^a; Z^g, \theta) $, and then fix $Z^a$ and sample $Z^g$ from $p(Z^g|X;Z^a,\theta)$ $\propto$ $p(X, Z^g;Z^a, \theta)$. At each sampling step, the latent vectors are updated as follows:
\begin{eqnarray}
Z_{\tau+1}^a  = Z_{\tau}^a + \frac{\delta^2}{2}\frac{\partial}{\partial Z^a}\log p(X, Z_{\tau}^a; Z_{\tau}^g, \theta) +
{\delta} {\cal E}_{\tau}^a \nonumber\\
Z_{\tau+1}^g  = Z_{\tau}^g + \frac{\delta^2}{2}\frac{\partial}{\partial Z^g}\log p(X, Z_{\tau}^g;Z_{\tau}^a, \theta) +
{\delta} {\cal E}_{\tau}^g
\label{eq:Langevin}
\end{eqnarray}
where $\tau$ is the number of steps in the Langevin sampling, ${\cal E}_{\tau}^a$, ${\cal E}_{\tau}^g$ are independent standard Gaussian noise to prevent the sampling from being trapped in local modes, and $\delta$ is the step size. The complete-data log-likelihood can be evaluated by
\begin{eqnarray}
\log p(X, Z^a;Z^g, \theta) = \log \left[p(Z^a) p(X|Z^a,Z^g, \theta) \right]\nonumber\\
= -\frac{1}{2\sigma^2} \|X - F(Z^a,Z^g; \theta)\|^2 -\frac{1}{2} \|Z^a\|^2 + C_1 \nonumber\\
\log p(X, Z^g;Z^a, \theta) = \log \left[p(Z^g) p(X|Z^a,Z^g, \theta) \right]\nonumber\\
= -\frac{1}{2\sigma^2} \|X - F(Z^a,Z^g; \theta)\|^2 -\frac{1}{2} \|Z^g\|^2 + C_2
\end{eqnarray}
where $C_1$ and $C_2$ are normalizing constants. It can be shown that, given sufficient  sampling steps, the sampled $Z^a$  and $Z^g$ follow their joint posterior distribution.

Obtaining fair samples from the posterior distribution by MCMC is highly computational consuming. In this paper, we run persistent sampling chains. That is, the MCMC sampling at each iteration starts from the sampled $Z^a$ and $Z^g$ in the previous iteration. The persistent update results in a chain that is long enough to sample from the posterior distribution, and vastly reduces the computational burden of the MCMC sampling. The convergence of stochastic gradient descent based on persistent MCMC has been studied in \cite{younes1999convergence}.

For each training example $X_i$, we run the Langevin dynamics following Eq.(\ref{eq:Langevin}) to get the corresponding posterior samples $Z_i^a$ and $Z_i^g$. The samples are then used for computing the gradients over parameters as shown in Eq.(\ref{eq:expectposterior}). More precisely, the gradient of log-likelihood over $\theta$ is estimated by Monte Carlo approximation:
\begin{align}
\frac{\partial}{\partial \theta}L&(\theta)  \approx \frac{1}{N}\sum_{i=1}^N \frac{\partial}{\partial \theta} \log p(X_i, Z^a_i,Z^g_i; \theta)\nonumber\\
&=\frac{1}{N}\sum_{i=1}^N \frac{1}{\sigma^2} (X_i-F(Z^a_i,Z^g_i; \theta))\frac{\partial}{\partial \theta}F(Z^a_i,Z^g_i; \theta).
\label{eq:weight}
\end{align}

The whole algorithm iterates through two steps: (1) inferential step which infers the latent vectors by Langevin dynamics, and (2) learning step which learns the network parameters $\theta$ by stochastic gradient descent. Gradient computations in both steps are powered by back-propagation. Algorithm $\ref{alg:1}$ describes the details of the learning and inference algorithm.

\begin{algorithm}[t]
\caption{Learning and inference algorithm}
\label{alg:1}
\begin{algorithmic}
\REQUIRE   \quad \\ (1) training examples $\{X_i \in \mathbb{R}^{D_x \times D_y \times 3}, i = 1, \dots, N\}$\\(2) number of  Langevin steps $l$\\(3) number of learning iterations $K$
\ENSURE \quad \\ (1) learned parameters $\theta$\\(2) inferred latent vectors  $\{Z^a_i, Z^g_i, i=1,\dots,N\}$\\~\\

\STATE 1:  Let $k \leftarrow 0$, initialize $\theta$.
\STATE 2:  Initialize $\{Z^a_i, Z^g_i , i=1,\dots,N\}$
\REPEAT
\STATE 3: \textbf{Inference back-propagation:} For each $i$, run $l$ steps of Langevin dynamics to alternately sample $Z^a_i$ from $p(Z^a_i|X_i;Z^g_i,\theta)$, while fixing $Z^g_i$; and sample $Z^g_i$ from $p(Z^g_i|X_i;Z^a_i,\theta)$, while fixing $Z^a_i$. Starting from the current $Z^a_i$ and $Z^g_i$, each step follows Eq.(\ref{eq:Langevin}). \\
\STATE 4: \textbf{Learning back-propagation:} Update $\theta_{k+1} \leftarrow \theta_k+\eta_k L^{\prime}(\theta_k)$, with learning rate $\eta_k$, where $L^{\prime}(\theta_k)$ is computed according to Eq.(\ref{eq:weight}).\\
\STATE 5: Let $k \leftarrow k+1$
\UNTIL{$k=K$}
\end{algorithmic}
\end{algorithm}
Besides the above learning and inference method, the proposed model can also be learned by VAE \cite{kingma2013auto} with an extra inference network to infer $(Z^a, Z^g)$, or learned by GAN \cite{goodfellow2014generative} with an extra discriminator network. In this work, we use the current learning and inference algorithm mainly for the sake of simplicity, so that we do not need to recruit to extra networks.
\section{Dynamic deformable model and learning algorithm}
For the spatial-temporal process in video, we propose a dynamic deformable model to disentangle the appearance and geometric information. Specifically, a sequence of appearance latent factors $\{Z^a_t, t=0,\dots, T\}$, and a sequence of geometric latent factors $\{Z^g_t, =0,\dots, T\}$ are fed into two generator networks, which are combined by a warping function to produce the observed video sequence $\mathbf{X}=\{X_0,\dots,X_T\}$. Suppose $Z^a_0 \in \mathbb{R}^{d_a}$ and $Z^g_0 \in \mathbb{R}^{d_g}$ are the appearance and geometric latent factors of the first frame, then
\begin{eqnarray}
Z^a_t &=& Z^a_0 + s^a_t,\nonumber\\
Z^g_t &=& Z^g_0 + s^g_t,
\label{eq:latentsum}
 \end{eqnarray}
where $s^a_t \in \mathbb{R}^{d_a}$ and $s^g_t \in \mathbb{R}^{d_g}$ are the hidden state vectors that capture the dynamic relation among the video sequence data, $s^a_0=s^g_0=0$. Inspired by \cite{doretto2003dynamic} which employs linear auto-regressive model to model the dynamic textures, we introduce non-linear auto-regressive models to model the transition between the hidden state vectors  $s^a_t$ and $s^g_t$:
\begin{eqnarray}
s_{t+1}^a  &=& f(s_t^a,\xi_{t+1}^a; \alpha) \nonumber\\
s_{t+1}^g  &=& f(s_t^g,\xi_{t+1}^g; \beta)
\label{eq:transition}
\end{eqnarray}
where $\xi_{t+1}^a$ and $\xi_{t+1}^g$ are independent Gaussian noise vectors, that encode the randomness in the transition from $s_t^a$ to $s_{t+1}^a$ and $s_t^g$ to $s_{t+1}^g$. $f(\cdot;\alpha)$ and $f(\cdot;\beta)$ are two feedforward neural networks or multi-layer perceptrons, where $\alpha$ and $\beta$ denote the weight and bias parameters of the two networks.

The dynamic deformable generator model can be expressed as
\begin{eqnarray}
X_t &= F_w(F_a(Z^a_t;\theta_a),F_g(Z^g_t;\theta_g)) + \epsilon_t
\label{eq:dynamodel}
\end{eqnarray}
where $F_w$ is the warping function as described in Section 3.2. $F_a(\cdot;\theta_a)$ and $F_g(\cdot;\theta_g)$ are the emission models, more specially, the appearance generator and geometric generator, $\theta_a$ and $\theta_g$ denote the weight and
bias parameters of the two generator networks. $\epsilon_t \sim {\rm N}(0, \sigma^2 I_D)$  is the independent residual error.

The proposed dynamic deformable generator model can be learned by alternating back-propagation for two sequences of latent vectors without introducing assisting inference network.
Our method is motivated by a maximum likelihood learning algorithm for time series data. Specifically, let $\boldsymbol{\theta}=\{\theta_a,\theta_g,\alpha,\beta\}$ consists of all the network parameters to be learned. Let $\mathbf{Z^a}= \{Z^a_0, \xi_{1}^a, \dots, \xi_{T}^a\}$ denotes the appearance related latent vectors and $\mathbf{Z^g}= \{Z^g_0, \xi_{1}^g, \dots, \xi_{T}^g\}$ be the geometric related latent  vectors. Both $\mathbf{Z^a}$ and $\mathbf{Z^g}$ can be inferred from the observed video sequences $\mathbf{X}$.
We can formulate
\begin{eqnarray}
\mathbf{X} &= F(\mathbf{Z^a},\mathbf{Z^g}; \boldsymbol{\theta}) +\boldsymbol{\epsilon}
\label{eq:dynamodelall}
\end{eqnarray}
where $F(\cdot,\cdot; \boldsymbol{\theta})$ composes $F_a$, $F_g$ and $f$ over time, and $\boldsymbol{ \epsilon}=\{\epsilon_t,t=0,\dots,T\}$ denotes the observation errors.

The proposed model  can be trained by maximizing the log-likelihood on the training video dataset $\{\mathbf{X_i},i=1,\dots,N\}$,
\begin{eqnarray}
L(\boldsymbol{\theta}) &=& \frac{1}{N}\sum_{i=1}^N\log p(\mathbf{X_i}; \boldsymbol{\theta})  \nonumber \\
&=& \frac{1}{N} \sum_{i=1}^N \log \int p(\mathbf{X_i}, \mathbf{Z^a_i}, \mathbf{Z^g_i}; \boldsymbol{\theta})d\mathbf{Z^a_i}d\mathbf{Z^g_i},
\label{eq:llevideo}
\end{eqnarray}
The gradient of the observed-data log-likelihood $L(\boldsymbol{\theta})$ can be evaluated similarly as in Eq. (\ref{eq:expectposterior}) of Section 3.3,
\begin{eqnarray}
\frac{\partial}{\partial \boldsymbol{\theta}} \log p(\mathbf{X}; \boldsymbol{\theta})=\E_{p( \mathbf{Z^a},\mathbf{Z^g}|\mathbf{X}; \boldsymbol{\theta})} \left[\frac{\partial}{\partial \boldsymbol{\theta}} \log p(\mathbf{X}, \mathbf{Z^a},\mathbf{Z^g}; \boldsymbol{\theta}) \right]
\label{eq:expectposteriorvideo}
\end{eqnarray}
We employ the Monte Carlo average to approximate the above expectation. Specifically, we sample from the posterior distribution $p( \mathbf{Z^a},\mathbf{Z^g}|\mathbf{X}; \boldsymbol{\theta})$ alternately by Langevin dynamics to infer the group of latent vectors $\mathbf{Z^a}$ and $\mathbf{Z^g}$.
We fix $\mathbf{Z^g}$ and sample $\mathbf{Z^a}$ from $p(\mathbf{Z^a}|\mathbf{X};\mathbf{Z^g},\boldsymbol{\theta})$ $\propto $ $p(\mathbf{X}, \mathbf{Z^a}; \mathbf{Z^g}, \boldsymbol{\theta}) $, and then fix $\mathbf{Z^a}$ and sample $\mathbf{Z^g}$ from $p(\mathbf{Z^g}|\mathbf{X};\mathbf{Z^a},\boldsymbol{\theta})$ $\propto$ $p(\mathbf{X}, \mathbf{Z^g};\mathbf{Z^a}, \boldsymbol{\theta})$. At each sampling step, the group of latent vectors are updated as follows:
\begin{eqnarray}
\mathbf{Z_{\tau+1}^a}  = \mathbf{Z_{\tau}^a} + \frac{\delta^2}{2}\frac{\partial}{\partial \mathbf{Z^a}}\log p(\mathbf{X}, \mathbf{Z_{\tau}^a} ; \mathbf{Z_{\tau}^g} , \boldsymbol{\theta}) +
{\delta} \boldsymbol{{\cal E}_{\tau}^a} \nonumber\\
\mathbf{Z_{\tau+1}^g}  = \mathbf{Z_{\tau}^g} + \frac{\delta^2}{2}\frac{\partial}{\partial \mathbf{Z^g}}\log p(\mathbf{X}, \mathbf{Z_{\tau}^g} ; \mathbf{Z_{\tau}^a} , \boldsymbol{\theta}) +
{\delta} \boldsymbol{{\cal E}_{\tau}^g}
\label{eq:Langevinvideo}
\end{eqnarray}
where $\tau$ is the number of steps in the Langevin sampling (not to be confused with the time step of the  dynamic model,t), $\boldsymbol{{\cal E}_{\tau}^a}$, $\boldsymbol{{\cal E}_{\tau}^g}$ are independent standard Gaussian noise to prevent the sampling from being trapped in local modes, and $\delta$ is the step size. $\mathbf{Z_{\tau}^a} = \{Z^a_{\tau,0}, \xi_{\tau,1}^a, \dots, \xi_{\tau,T}^a\}$ and $\mathbf{Z_{\tau}^g} = \{Z^g_{\tau,0}, \xi_{\tau,1}^g, \dots, \xi_{\tau,T}^g\}$ denote all the sampled appearance and geometric latent vectors at time step $\tau$.

Let $p(\mathbf{X}|\mathbf{Z^a},\mathbf{Z^g}, \boldsymbol{\theta})$ $\sim$ ${\rm N}( F(\mathbf{Z^a},\mathbf{Z^g}; \boldsymbol{\theta}), \sigma^2 \mathbf{I})$, where $\mathbf{I}$ is the identity matrix whose dimension matches that of $\mathbf{X}$. Let $p(\mathbf{Z^a})$ $\sim$ ${\rm N}(0, \mathbf{I_a}) $ and $p(\mathbf{Z^g})$ $\sim$ ${\rm N}(0, \mathbf{I_g}) $be the prior distribution of $\mathbf{Z^a}$ and $\mathbf{Z^g}$, $\mathbf{I_a}$ and $\mathbf{I_g}$ are the identity matrices whose dimensions match that of $\mathbf{Z^a}$ and $\mathbf{Z^g}$. The complete-data log-likelihood can be evaluated by (assuming $\sigma^2=1$, and up to an additive constant)
\begin{align}
&\log p(\mathbf{X}, \mathbf{Z^a};\mathbf{Z^g}, \boldsymbol{\theta}) = \log \left[p(\mathbf{Z^a}) p(\mathbf{X}|\mathbf{Z^a},\mathbf{Z^g}, \boldsymbol{\theta}) \right]\nonumber\\
&= -\frac{1}{2}( \|\mathbf{X} - F(\mathbf{Z^a},\mathbf{Z^g}; \boldsymbol{\theta})\|^2 + \|\mathbf{Z^a}\|^2) \nonumber\\
&=-\frac{1}{2}(\sum_{t=0}^T  \|X_t - F_w(F_a(Z^a_t;\theta_a),F_g(Z^g_t;\theta_g))  \|^2 +\nonumber\\
&\qquad \quad \|Z_0^a\|^2 + \sum_{t=1}^T \|\xi_{t}^a\|^2)\nonumber\\
&\log p(\mathbf{X}, \mathbf{Z^g};\mathbf{Z^a}, \boldsymbol{\theta}) = \log \left[p(\mathbf{Z^g}) p(\mathbf{X}|\mathbf{Z^a},\mathbf{Z^g}, \boldsymbol{\theta}) \right]\nonumber\\
&= -\frac{1}{2}( \|\mathbf{X} - F(\mathbf{Z^a},\mathbf{Z^g}; \boldsymbol{\theta})\|^2 + \|\mathbf{Z^g}\|^2) \nonumber\\
&=-\frac{1}{2}(\sum_{t=0}^T  \|X_t - F_w(F_a(Z^a_t;\theta_a),F_g(Z^g_t;\theta_g))  \|^2 +\nonumber\\
&\qquad \quad \|Z_0^g\|^2 + \sum_{t=1}^T \|\xi_{t}^g\|^2)
\label{eq:cllevideo}
\end{align}
To infer the detailed components in $\mathbf{Z^a}= \{Z^a_0, \xi_{1}^a, \dots, \xi_{T}^a\}$ and $\mathbf{Z^g}= \{Z^g_0, \xi_{1}^g, \dots, \xi_{T}^g\}$, for any fixed time point $t_0$,we have
\begin{align}
&\frac{\partial}{\partial \xi^a_{t_0}}\log p(\mathbf{X}, \mathbf{Z_{\tau}^a} ; \mathbf{Z_{\tau}^g} , \boldsymbol{\theta}) \nonumber\\
&=\sum_{t={t_0}}^T  (X_t - F_w(F_a(Z^a_t;\theta_a),F_g(Z^g_t;\theta_g))) \times \nonumber\\
&\quad \frac{\partial  F_w(F_a(Z^a_t;\theta_a),F_g(Z^g_t;\theta_g)) }{\partial F_a(Z^a_t;\theta_a)} \frac{\partial F_a(Z^a_t;\theta_a) }{\partial Z^a_t}\frac{\partial s^a_t}{\partial \xi^a_{t_0}} - \xi^a_{t_0} \nonumber
\end{align}

\begin{align}
&\frac{\partial}{\partial Z^a_0}\log p(\mathbf{X}, \mathbf{Z_{\tau}^a} ; \mathbf{Z_{\tau}^g} , \boldsymbol{\theta}) \nonumber\\
&=\sum_{t=0}^T  (X_t - F_w(F_a(Z^a_t;\theta_a),F_g(Z^g_t;\theta_g))) \times \nonumber\\
&\quad \frac{\partial  F_w(F_a(Z^a_t;\theta_a),F_g(Z^g_t;\theta_g)) }{\partial F_a(Z^a_t;\theta_a)} \frac{\partial F_a(Z^a_t;\theta_a) }{\partial Z^a_t} - Z^a_0 \nonumber
\end{align}

\begin{align}
&\frac{\partial}{\partial \xi^g_{t_0}}\log p(\mathbf{X}, \mathbf{Z_{\tau}^g} ; \mathbf{Z_{\tau}^a} , \boldsymbol{\theta}) \nonumber\\
&=\sum_{t={t_0}}^T  (X_t - F_w(F_a(Z^a_t;\theta_a),F_g(Z^g_t;\theta_g))) \times \nonumber\\
&\quad \frac{\partial  F_w(F_a(Z^a_t;\theta_a),F_g(Z^g_t;\theta_g)) }{\partial F_g(Z^g_t;\theta_g)} \frac{\partial F_g(Z^g_t;\theta_g) }{\partial Z^g_t}\frac{\partial s^g_t}{\partial \xi^g_{t_0}} - \xi^g_{t_0} \nonumber
\end{align}

\begin{align}
&\frac{\partial}{\partial Z^g_0}\log p(\mathbf{X}, \mathbf{Z_{\tau}^g} ; \mathbf{Z_{\tau}^a} , \boldsymbol{\theta}) \nonumber\\
&=\sum_{t=0}^T  (X_t - F_w(F_a(Z^a_t;\theta_a),F_g(Z^g_t;\theta_g))) \times \nonumber\\
&\quad \frac{\partial  F_w(F_a(Z^a_t;\theta_a),F_g(Z^g_t;\theta_g)) }{\partial F_g(Z^g_t;\theta_a)} \frac{\partial F_g(Z^g_t;\theta_a) }{\partial Z^g_t} - Z^g_0
\label{eq:compZ}
\end{align}
where $\frac{\partial s^a_t}{\partial \xi^a_{t_0}}$ and $\frac{\partial s^g_t}{\partial \xi^g_{t_0}}$ can be computed recursively.

The learning algorithm iterates through two steps: (1) inference step: infers the two group of latent vectors $\mathbf{Z^a}$  and $\mathbf{Z^g}$ through the Langevin dynamics, when the current $\boldsymbol{\theta}$ is given, according to Eq.(\ref{eq:Langevinvideo}$\sim$ \ref{eq:compZ}). (2) Learning step: update the network parameters $\boldsymbol{\theta}$  by stochastic gradient descent, when the group of samples $\mathbf{Z^a}$  and $\mathbf{Z^g}$ are given, according to Eq. (\ref{eq:llevideo}) and (\ref{eq:expectposteriorvideo}), more precisely,
\begin{eqnarray}
\frac{\partial}{\partial \boldsymbol{\theta}}L(\boldsymbol{\theta}) \approx \frac{1}{N}\sum_{i=1}^N \frac{\partial}{\partial \boldsymbol{\theta}} \log p(\mathbf{X_i}, \mathbf{Z^a_i},\mathbf{Z^g_i}; \boldsymbol{\theta})\nonumber\\
=\frac{1}{N}\sum_{i=1}^N \frac{1}{\sigma^2} (\mathbf{X_i}-F(\mathbf{Z^a_i},\mathbf{Z^g_i}; \boldsymbol{\theta}))\frac{\partial}{\partial \boldsymbol{\theta}}F(\mathbf{Z^a_i},\mathbf{Z^g_i}; \boldsymbol{\theta}).
\label{eq:weightvideo}
\end{eqnarray}
where $\boldsymbol{\theta}=\{\theta_a,\theta_g,\alpha,\beta\}$, i indexes the video sequence in the training set. For the i-th video sequence, the derivative with respect to the components of  $\boldsymbol{\theta}$ are (for convenience and simplicity, omit the index i, assuming $\sigma^2=1$)
\begin{align}
&\frac{\partial L}{\partial \theta_a}= \sum_{t=0}^T  (X_t - F_w(F_a(Z^a_t;\theta_a),F_g(Z^g_t;\theta_g))) \times \nonumber\\
&\qquad \quad \frac{\partial  F_w(F_a(Z^a_t;\theta_a),F_g(Z^g_t;\theta_g)) }{\partial F_a(Z^a_t;\theta_a)} \frac{\partial F_a(Z^a_t;\theta_a) }{\partial \theta_a} \nonumber
\end{align}
\begin{align}
&\frac{\partial L}{\partial \theta_g}= \sum_{t=0}^T  (X_t - F_w(F_a(Z^a_t;\theta_a),F_g(Z^g_t;\theta_g))) \times \nonumber\\
&\qquad \quad \frac{\partial  F_w(F_a(Z^a_t;\theta_a),F_g(Z^g_t;\theta_g)) }{\partial F_g(Z^g_t;\theta_g)} \frac{\partial F_g(Z^g_t;\theta_g) }{\partial \theta_g} \nonumber
\end{align}
\begin{align}
&\frac{\partial L}{\partial \alpha}= \sum_{t=0}^T  (X_t - F_w(F_a(Z^a_t;\theta_a),F_g(Z^g_t;\theta_g))) \times \nonumber\\
&\qquad \quad \frac{\partial  F_w(F_a(Z^a_t;\theta_a),F_g(Z^g_t;\theta_g)) }{\partial F_a(Z^a_t;\theta_a)} \frac{\partial F_a(Z^a_t;\theta_a) }{\partial Z^a_t}\frac{\partial s^a_t}{\partial \alpha} \nonumber
\end{align}
\begin{align}
&\frac{\partial L}{\partial \beta}= \sum_{t=0}^T  (X_t - F_w(F_a(Z^a_t;\theta_a),F_g(Z^g_t;\theta_g))) \times \nonumber\\
&\qquad \quad \frac{\partial  F_w(F_a(Z^a_t;\theta_a),F_g(Z^g_t;\theta_g)) }{\partial F_g(Z^g_t;\theta_g)} \frac{\partial F_g(Z^g_t;\theta_g) }{\partial Z^g_t}\frac{\partial s^g_t}{\partial \beta}
\label{eq:comptheta}
\end{align}
where $\frac{\partial s^a_t}{\partial \alpha}$ and $\frac{\partial s^g_t}{\partial \beta} $ can again be computed recursively. The gradient computations in both the inferential step and the learning step are powered by back-propagation through time (BPTT). Algorithm \ref{alg:2} describes the details of the learning and inference algorithm powered by BPTT for the video sequences.

It is worth to note that, although the variational inference is convenient to be employed for learning a generative model, for this dynamic deformable model, it is difficult to design the inference models to infer the sequence of the group of latent vectors $\mathbf{Z^a}= \{Z^a_{0}, \xi_{1}^a, \dots, \xi_{T}^a\}$ and $\mathbf{Z^g}= \{Z^g_{0}, \xi_{1}^g, \dots, \xi_{T}^g\}$ from the video sequence $\mathbf{X}=\{X_0,\dots,X_T\}$.  In contrast, the proposed learning and inference algorithm does not need to design and learn such an extra inference model, and is easy to be implemented. Specifically, we directly sample from the posterior distribution $p( \mathbf{Z^a},\mathbf{Z^g}|\mathbf{X}; \boldsymbol{\theta})$ to implement the inference step in our dynamic deformable model, which is powered by BPTT. Moreover, the proposed learning method directly aims at maximum likelihood, while variational inference targets at maximizing a lower bound.
\begin{algorithm}[t]
\caption{Learning and inference algorithm for video sequences}
\label{alg:2}
\begin{algorithmic}
\REQUIRE   \quad \\ (1) training examples $\{\mathbf{X_i} \in \mathbb{R}^{D_x \times D_y \times 3 \times (T+1)}, i = 1, \dots, N\}$\\(2) number of  Langevin steps $l$\\(3) number of learning iterations $K$
\ENSURE \quad \\ (1) learned parameters $\boldsymbol{\theta}=\{\theta_a,\theta_g,\alpha,\beta\}$\\(2) inferred group of latent vectors,\\ $\mathbf{Z^a_i}= \{Z^a_{0,i}, \xi_{1,i}^a, \dots, \xi_{T,i}^a\}$ and $\mathbf{Z^g_i}= \{Z^g_{0,i}, \xi_{1,i}^g, \dots, \xi_{T,i}^g\}$\\~\\

\STATE 1:  Let $k \leftarrow 0$, initialize $\boldsymbol{\theta}=\{\theta_a,\theta_g,\alpha,\beta\}$.
\STATE 2:  Initialize $\{\mathbf{Z^a_i}, \mathbf{Z^g_i} , i=1,\dots,N\}$
\REPEAT
\STATE 3: \textbf{Inference back-propagation through time:} For each $i$, run $l$ steps of Langevin dynamics to alternately sample $\mathbf{Z^a_i}$ from $p(\mathbf{Z^a_i}|\mathbf{X};\mathbf{Z^g_i},\boldsymbol{\theta})$, while fixing $\mathbf{Z^g_i}$; and sample $\mathbf{Z^g_i}$ from $p(\mathbf{Z^g_i}|\mathbf{X};\mathbf{Z^a_i},\boldsymbol{\theta})$, while fixing $\mathbf{Z^a_i}$. Starting from the current $\mathbf{Z^a_i}$ and $\mathbf{Z^g_i}$, each step follows Eq.(\ref{eq:Langevinvideo}$\sim$ \ref{eq:compZ}). \\
\STATE 4: \textbf{Learning back-propagation through time:} Update $\boldsymbol{\theta^{k+1}} \leftarrow \boldsymbol{\theta^k}+\eta_k L^{\prime}(\boldsymbol{\theta^k})$, with learning rate $\eta_k$, where $L^{\prime}(\boldsymbol{\theta^k}=\{\theta_a^k,\theta_g^k,\alpha^k,\beta^k\})$ is computed according to Eq.(\ref{eq:weightvideo},\ref{eq:comptheta}).\\
\STATE 5: Let $k \leftarrow k+1$
\UNTIL{$k=K$}
\end{algorithmic}
\end{algorithm}
\section{Experiments}
In this section, we design 6 groups of experiments to demonstrate that our proposed deformable generator framework consistently disentangles the appearance and geometric information. The parameters and architectures of the deformable generator network are summarized in subsection 5.7. In the following experiments, in each row we visualize the generated samples by varying a certain unit of the latent factors within the range $[-\gamma, \gamma]$, where we set $\gamma$ to be $10$. The code and results can be found at the project page~\footnote{\url{https://andyxingxl.github.io/Deformable-generator/}}. We have implemented the algorithm with the MindSpore framework~\footnote{\url{https://www.mindspore.cn/en}}.

\subsection{Experiment 1: Learning the disentangled basis functions for appearance and geometry}
\label{ex1}
\begin{figure*}[t!]
\begin{center}
\includegraphics[width=1.7\columnwidth]{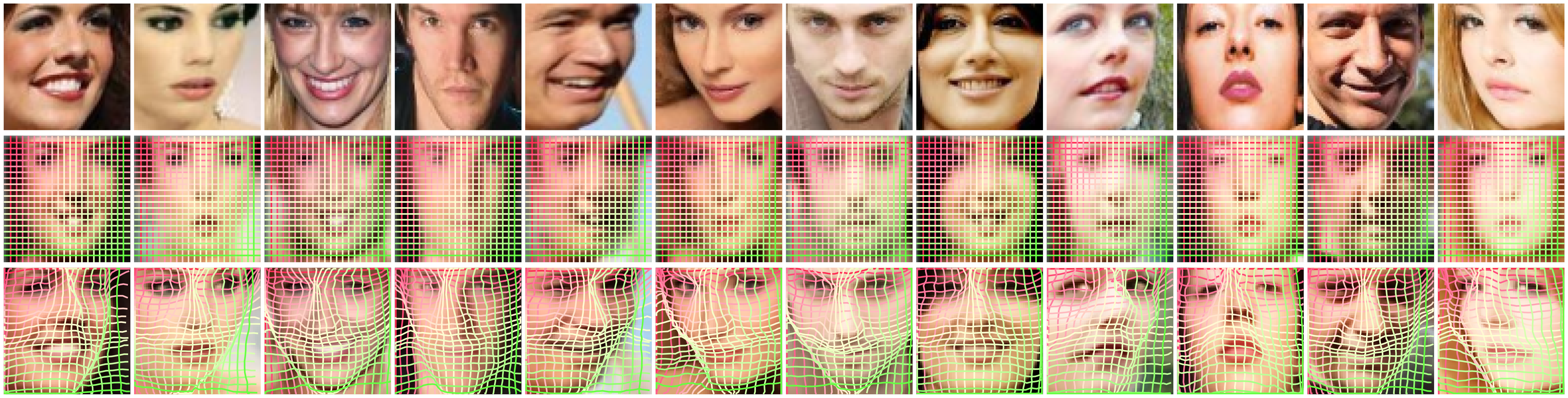}\\
\caption{Example training images from CelebA are illustrated at the first row. The training set contains 10000 images from CelebA, and they are cropped to $64 \times 64$ pixels by the OpenFace. These faces have different colors, illuminations, identities, viewing angles, shapes, and expressions. The second row shows the output of the appearance generator overlapped with  the canonical grid. The third row demonstrates the deformation fields which is the output of the geometric generator. The deformation fields are visualized by the deformed grids overlaid on the reconstructed images.}
\label{fig:exampleceleb}
\end{center}
\end{figure*}

To study the performance of the proposed method in disentangling the appearance and geometric information, we first investigate the appearance basis functions and the geometric basis functions of the learned model. We train the deformable generator on 10,000 face images randomly sampled from CelebA dataset \cite{liu2015faceattributes}. Some examples in CelebA are shown in Figure \ref{fig:exampleceleb}, which are processed by the OpenFace \cite{amos2016openface} and further cropped to $64 \times 64$ pixels. To better understand how our model works, we show the output of the appearance generator overlaid by a canonical grid in the second row of figure \ref{fig:exampleceleb}. The canonical faces in the front view are learned by the appearance generator. By warping the output of the appearance generator with the deformation fields generated by the geometric generator, we obtain the final reconstructed images, which are shown in the third row of figure \ref{fig:exampleceleb}. The deformation fields, which are the output of the geometric generator, are illustrated by the deformed grids overlaid on the reconstruct images.

\begin{figure}[ht]
\begin{center}
 \includegraphics[width=\columnwidth]{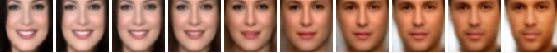}\\
 \includegraphics[width=\columnwidth]{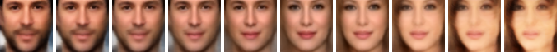}\\
 \includegraphics[width=\columnwidth]{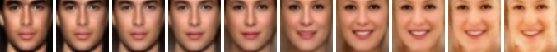}\\
 \includegraphics[width=\columnwidth]{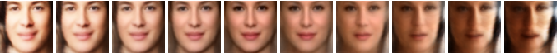}
\caption{Typical appearance basis functions, visualized by the generated images that interpolate the appearance latent factors along the basis functions. Each dimension of the appearance latent factors encodes appearance information such as color, illumination and gender. In the fist line, the color of background and the gender change. In the second line, the moustache of the man and the hair of the woman vary. In the third line, the skin color changes from dark to white. In the fourth line, the illumination lighting changes from the left-side of the face to the right-side of the face.}
\label{fig:app}
\end{center}
\end{figure}

\begin{figure}[ht]
\begin{center}
 \includegraphics[width=\columnwidth]{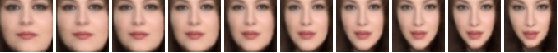}\\
 \includegraphics[width=\columnwidth]{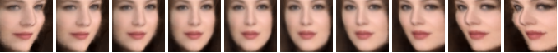}\\
 \includegraphics[width=\columnwidth]{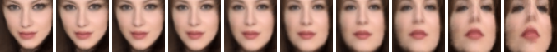}\\
 \includegraphics[width=\columnwidth]{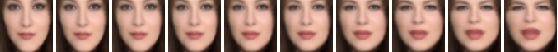}
\caption{Representative geometric basis functions, visualized by the generated images that interpolate the geometric latent factors along the basis functions. Each dimension of the geometric latent factors encodes fundamental geometric information such as shape and viewing angle. In the fist line, the shape of the face changes from fat to thin from left to the right. In the second line, the pose of the face varies from left to right. In the third line, from left to right, the vertical tilt of the face varies from downward to upward. In the fourth line, the face width changes from stretched to cramped.}
\label{fig:geo}
\end{center}
\end{figure}

The appearance and the geometric latent factors can be interpreted as the projection or reconstruction coefficients along the direction of the corresponding appearance and geometric basis functions. The appearance basis function of the learned model can be demonstrated by the generated images from the combinations of the appearance latent factors $Z^a$ and the geometric latent factors $Z^g$ as follows: (1) set the geometric latent factor $Z^g$ to zero, and (2) vary one dimension of the appearance variable $Z^a$ from $[-\gamma,\gamma]$ with a uniform step $\frac{2\gamma}{10}$, while set the other dimensions of $Z^a$ to zero. Some generated images are shown in figure \ref{fig:app}. Similarly, the geometric basis function of the learned model can be demonstrated as follows: (1) set $Z^a$ to be a fixed value, and (2) each time vary one dimension of the geometric latent factor $Z^g$ from $[-\gamma,\gamma]$ with a uniform step $\frac{2\gamma}{10}$, while keeping the other dimensions of $Z^g$ at zero. Some generated results are shown in figure \ref{fig:geo}.

As we can observe from figure \ref{fig:app}, (1) although the training faces from CelebA have different viewing angles, the appearance basis functions only encode front-view information, and (2) each dimension of the appearance latent vector encodes appearance information such as color, illumination and identity. For example, in the fist line of figure \ref{fig:app}, from left to right, the color of background varies from black to white, and the identity of the face changes from a women to a man. In the second line of figure \ref{fig:app}, the moustache of the man becomes thicker when the value of the corresponding dimension of $Z^a$ decreases, and the hair of the woman becomes denser when the value of the corresponding dimension of $Z^a$ increases. In the third line, from left to right, the skin color varies from dark to white, and in the fourth line, from left to right, the illumination lighting changes from the left-side of the face to the right-side of the face.

From  figure \ref{fig:geo}, we have the following interesting observations. (1) The geometric basis functions do not encode any appearance information. The color, illumination and identity are the same across these generated images. (2) Each dimension of the geometric latent vector encodes fundamental geometric information such as shape and viewing angle. For example, in the fist line of figure \ref{fig:geo}, the shape of the face changes from fat to thin from left to the right; in the second line, the pose of the face varies from left to right; in the third line, from left to right, the tilt of the face varies from downward to upward; and in the fourth line, the expression changes from stretched to cramped.

\begin{figure}[th]
\begin{center}
 \includegraphics[width=\columnwidth]{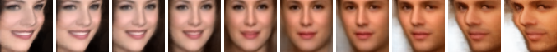}\\
 \includegraphics[width=\columnwidth]{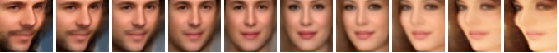}\\
 \includegraphics[width=\columnwidth]{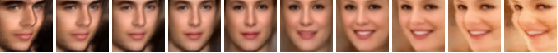}\\
\includegraphics[width=\columnwidth]{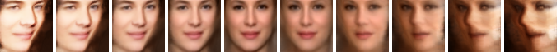}\\
 {\footnotesize(a) Rotation warping.} \\
  \includegraphics[width=\columnwidth]{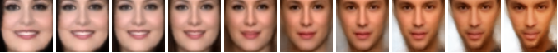}\\
 \includegraphics[width=\columnwidth]{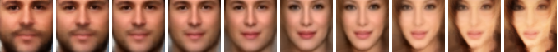}\\
 \includegraphics[width=\columnwidth]{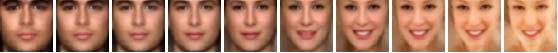}\\
 \includegraphics[width=\columnwidth]{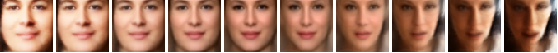}\\
 {\footnotesize(b) Shape warping.}

\caption{Applying the geometric basis functions, (a) rotation warping and (b) shape warping, learned by the geometric generator to the canonical faces generated by the appearance generator. Compared with figure \ref{fig:app}, only the pose information varies, and the identity information is kept in the process of warping.}
\label{fig:geoapplyrotate}
\end{center}
\end{figure}

From the results in figures \ref{fig:app} and \ref{fig:geo}, we find that the appearance and geometric information of face images have been disentangled effectively. Therefore, we can apply the geometric warping (e.g. geometric basis functions in figure \ref{fig:geo}) learned by the geometric generator to all the canonical faces (e.g. appearance basis functions in figure \ref{fig:app}) learned by the appearance generator. Figure \ref{fig:geoapplyrotate} demonstrates the effect of applying representative geometric basis functions to the appearance basis functions in figure \ref{fig:app}. Comparing figure \ref{fig:app} with figure \ref{fig:geoapplyrotate}, we find that the geometric basis functions which are corresponding to the rotation and shape warping operations do not modify the identity information of the canonical faces, which corroborates the disentangling power of the proposed deformable generator model.

\begin{figure}[h]
\begin{center}
 \includegraphics[width=\columnwidth]{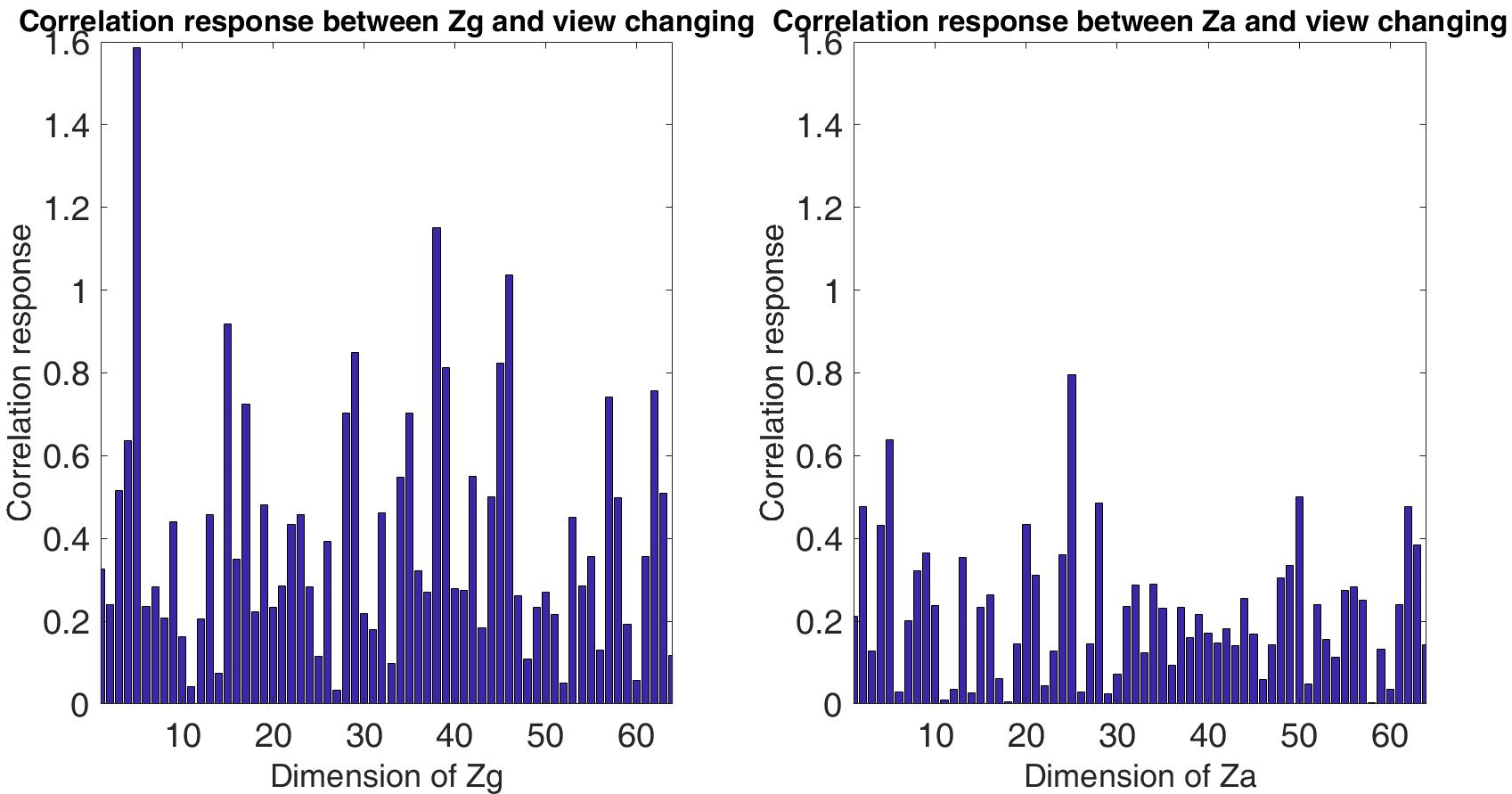}\\
\caption{Absolute value of covariance between each dimension of the geometric (or appearance) latent vectors and view variations for the face images from Multi-Pie. The left subfigure shows covariance with the geometric latent vector; the right subfigure shows covariance with the appearance latent vector.}
\label{fig:covarianceresponse}
\end{center}
\end{figure}

We next quantitatively study the covariance between the basis functions and input images with geometric variation. We use images with ground-truth geometric attributes, specifically the multi-view face images from the Multi-Pie dataset \cite{Gross2010}. The images consist of $5$ viewing angles $\{-30^\circ$, $-15^\circ$, $0^\circ$, $15^\circ$, $30^\circ\}$. For each viewing angle, we randomly sampled $100$ images, which are fed into the learned model to infer their geometric latent vector $Z^g$ and appearance latent vector $Z^a$. For each viewing angle $\theta$ , we compute the average $\bar{Z}^g_{\theta}$ and $\bar{Z}^a_{\theta}$ of the inferred latent vectors. For each dimension $i$ of $Z^g$, we construct a 5-dimensional vector $\bar{Z}^g(i)=[\bar{Z}^g_{-30^\circ}(i),\bar{Z}^g_{-15^\circ}(i),\bar{Z}^g_{0^\circ}(i),\bar{Z}^g_{15^\circ}(i),\bar{Z}^g_{30^\circ}(i)]$. Similarly, we construct a 5-dimensional vector $\bar{Z}^a(i)$ for each dimension of $Z^a$. We normalize the viewing angles vector $\theta = [-30,-15,0,15,30]$ to have unit norm. Finally, we compute the covariance between each dimension of the latent vectors ($Z^g, Z^a$) and input images with view variations as follows:
\begin{equation}
R^g_i = |\bar{Z}^g(i)^{\T}\theta|, \quad R^a_i = |\bar{Z}^a(i)^{\T}\theta|
\end{equation}
where $i$ denotes the $i$-th dimension of latent vector $Z^g$ or $Z^a$, and $|\cdot |$ denotes the absolute value. We summarize the the covariance $R^g$ and $R^a$ of the geometric and appearance latent vectors in figure \ref{fig:covarianceresponse}. $R^g$ tends to be much larger than $R^a$.

\begin{figure}[t]
\begin{center}
 \includegraphics[width=0.32\columnwidth]{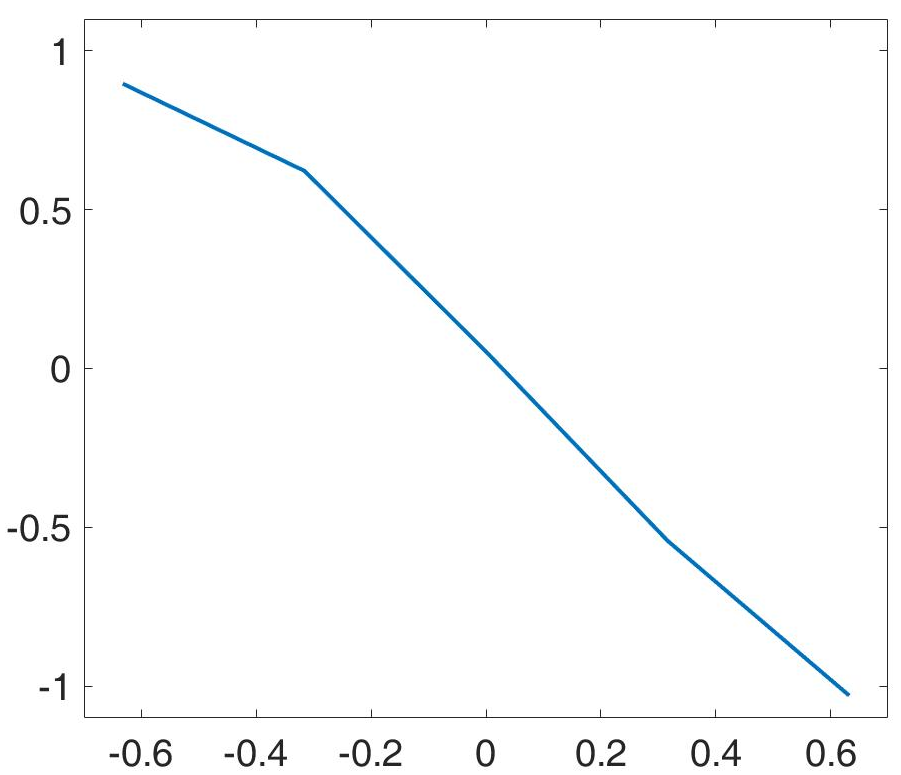}
 \includegraphics[width=0.32\columnwidth]{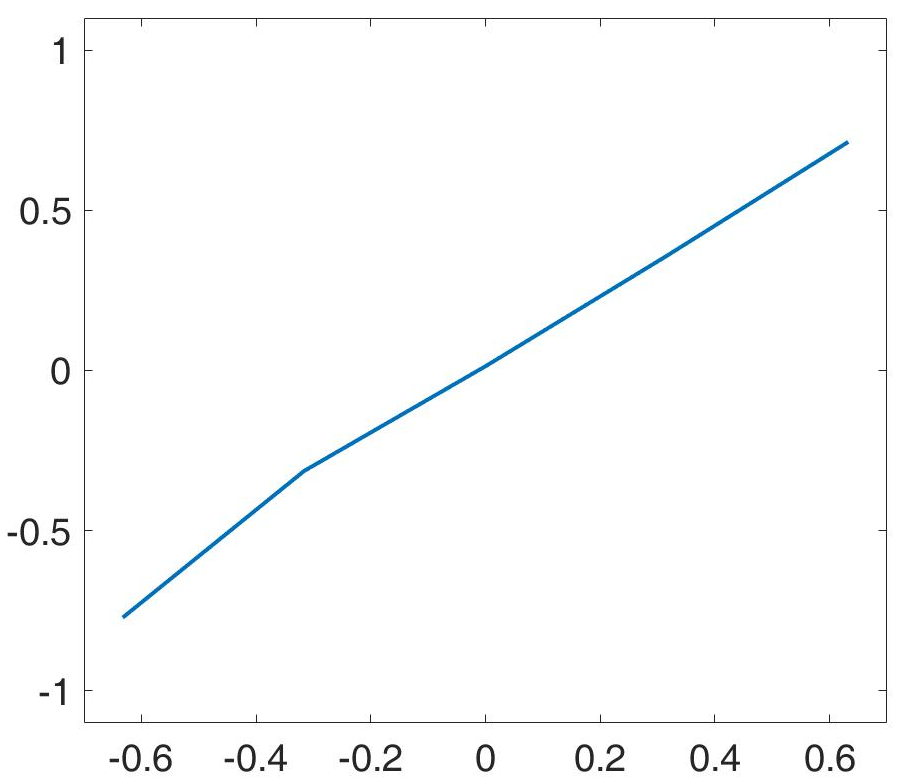}
 \includegraphics[width=0.32\columnwidth]{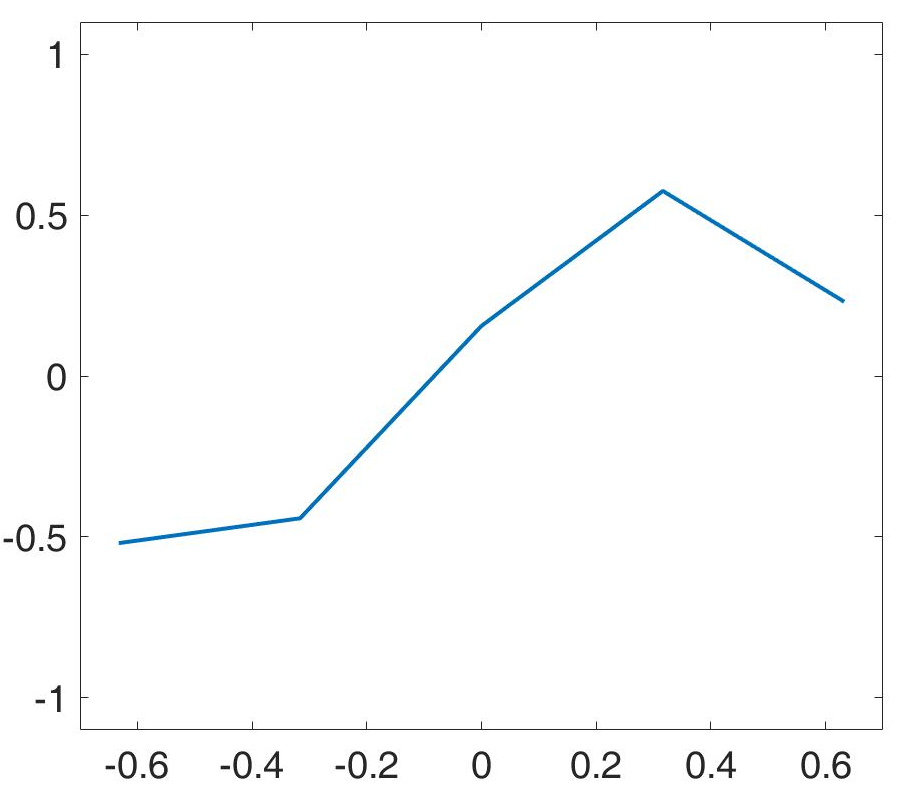}\\
 (a)\\
  \includegraphics[width=\columnwidth]{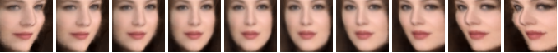}\\
 \includegraphics[width=\columnwidth]{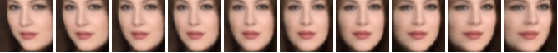}\\
 \includegraphics[width=\columnwidth]{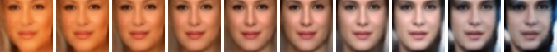}
 (b)
\caption{(a) Covariance relationship between the mean latent vector $\bar{Z}^g(i)$ (or $\bar{Z}^a(i)$) and viewing angles vector $\theta$. We choose two dimensions of $Z^g$ ($Z^g_5$ and $Z^g_{38}$, left and middle) with the largest covariance and one dimension of $Z^a$ with the largest covariance ($Z^a_{25}$, right). (b) Images generated by varying the values of the three dimensions in (a) respectively, while fixing the values of other dimensions to be zero. }
\label{fig:largestdimen}
\end{center}
\end{figure}

 Moreover, for the largest $R^g_i$ and largest $R^a_i$, we plot covariance relationship between the latent vector $\bar{Z}^g(i)$ (or $\bar{Z}^a(i)$) and viewing angles vector $\theta$ in figure \ref{fig:largestdimen}. As we can observe from the left and middle subfigures from figure \ref{fig:largestdimen}, the $\bar{Z}^g(i)$ corresponding to the two largest $R^g_i$ ($R^g_5$, $R^g_{38}$) is obviously inversely proportional or proportional to the change of viewing angle. However, as shown in the right subfigure, the $\bar{Z}^a(i)$ corresponding to the largest $R^a_i$ ($R^a_{25}$) does not have strong covariance with the change of viewing angle. We wish to point out that we should not expect $Z^a$ to encode the identity exclusively and $Z^g$ to encode the view exclusively, because different persons may have shape changes, and different views may have lighting or color changes.

Furthermore, we generate face images by varying the dimension of $Z^g$ corresponding to the two largest covariance responses from values $[-\gamma,+\gamma]$ with a uniform step $\frac{2\gamma}{10}$, while holding the other dimensions of $Z^g$ to zero. Similarly, we generate face images by varying the dimension of $Z^a$ corresponding to the largest covariance responses from values $[-\gamma,+\gamma]$ with a uniform step $\frac{2\gamma}{10}$, while holding the other dimensions of $Z^a$ to zero. The generated images are shown in figure \ref{fig:largestdimen}(b). We can make several important observations. (1) The variation of viewing angle in the first two rows is very obvious, and variation in the first row is larger than that the one in the second row. This is consistent with the fact that $R^g_5 > R^g_{38}$ and with the observation that the slope in the left subfigure of figure \ref{fig:largestdimen}(a) is steeper than that of the middle subfigure of figure \ref{fig:largestdimen}(a). (2) In the first row, the faces rotate from right to left, where $R^g_5$ is inversely proportional to the viewing angle. In the second row, the faces rotate from left to right, where $R^g_{38}$ is proportional to the viewing angle. (3) It is difficult to find obvious variation in viewing angle in the third row. These generated images further verify that the geometric generator of the proposed model mainly captures geometric variation, while the appearance generator is not sensitive to geometric variation.

\subsection{Experiment 2: Learning to transfer the appearance and geometric knowledge}
\begin{figure}[ht]
\begin{center}
 \includegraphics[width=\columnwidth]{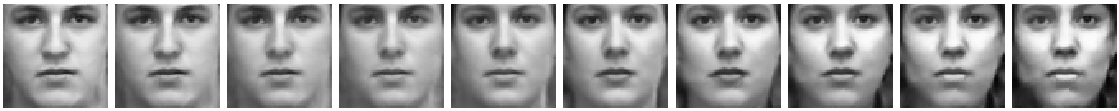}\\
 \includegraphics[width=\columnwidth]{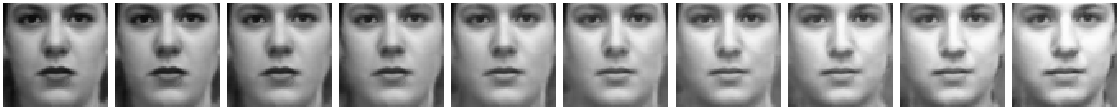}\\
 {\footnotesize (a) Interpolation of appearance latent factors.} \\
 \includegraphics[width=\columnwidth]{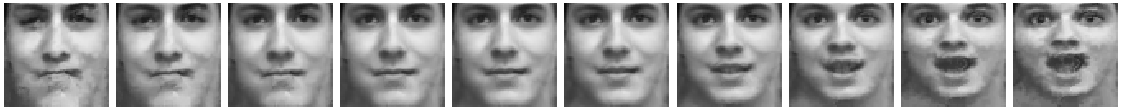}\\
 \includegraphics[width=\columnwidth]{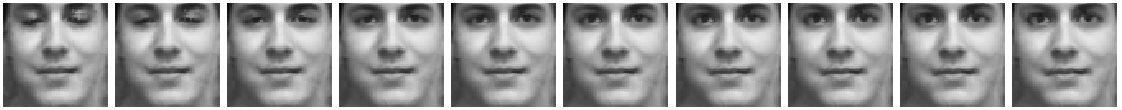}\\
 {\footnotesize(b) Interpolation of geometric latent factors.} \\
  \includegraphics[width=\columnwidth]{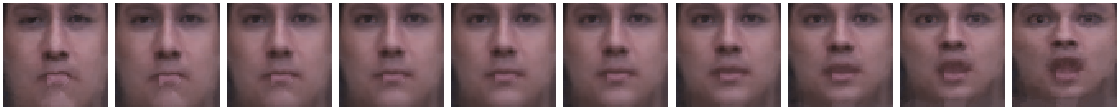}\\
 \includegraphics[width=\columnwidth]{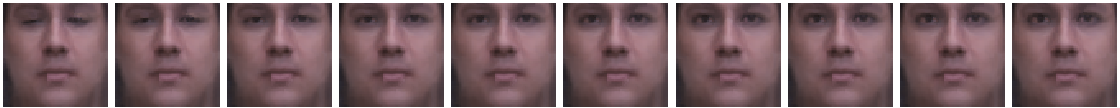}\\
 {\footnotesize(c) Transferring the expression in (b) to the face images in Multi-PIE dataset.}
\caption{Interpolation examples of (a) appearance basis functions  and (b) geometric basis functions. (c) Transferring the learned expression to the face images in Multi-PIE dataset.  }
\label{fig:appck}
\end{center}
\end{figure}

We can transfer the learned geometric knowledge from one dataset to another unseen dataset easily. We first train the proposed deformable generator model on the grey face expression dataset CK+ \cite{lucey2010extended}.
Following the same experimental protocol as the last subsection, we can investigate the interpolation along the appearance basis functions and the geometric basis functions. The disentangled results are shown in figure \ref{fig:appck}.  We do not use the labels of expressions provided by CK+ dataset in the learning. Although the dataset contains faces of different expressions, the learned appearance basis function usually encodes a neutral expression. The geometric basis function controls major variation in expression, but does not change the identity information.

Then, we try to transfer the learned geometric knowledge, such as expression, from CK+ to another color face dataset, Multi-Pie \cite{Gross2010}, by fine-turning the appearance generator on the target face dataset while fixing the parameters of the geometric generator.
Figure \ref{fig:appck} (c) shows the result of transferring the expressions of \ref{fig:appck} (b) into the faces of Multi-Pie. The expressions from the gray faces of CK+ have been transferred into the color faces of Multi-Pie.

Furthermore, we quantitatively study the power of the proposed deformable generator model to transfer the geometric knowledge learned from one dataset into another unseen dataset. Specifically, given 1000 front-view faces from the Multi-Pie dataset \cite{Gross2010}, we can fine-tune the appearance generator's parameters while fixing the geometric generator's parameters, which are learned from the CelebA dataset. Then we can reconstruct unseen images that have various viewpoints. In order to quantitatively evaluate the geometric knowledge transfer ability of our model, we compute the reconstruction error on 5000 unseen images from Multi-Pie for the views $\{-30^\circ$, $-15^\circ$, $0^\circ$, $15^\circ$, $30^\circ\}$, with 1000 faces for each view. We compare the proposed model with the state-of-art generative models, such as VAE \cite{kingma2013auto,burgess2018understanding} and ABP \cite{han2017alternating}. For fair comparison, we first train the original non-deformable VAE and ABP models with the same CelebA training set of 10,000 faces, and then fine-tune them on the 1000 front-view faces from the Multi-Pie dataset. We perform 10 independent runs and report the  mean square reconstruction error per image and standard derivation over the 10 trials  for each method under different views as shown in table \ref{table:reconstruction}. Deformable generator network obtains the lowest reconstruction error.  When the testing images are from the view closing to that of the training images, all the three methods can obtain small reconstruction errors. When various views of the testing images are included, deformable generator network obtains obviously smaller reconstruction error. Our model benefits from the transferred geometric knowledge learned from the CelebA dataset, while both the non-deformable VAE and ABP models cannot efficiently learn or transfer purely geometric information.

\begin{table}[t]
\caption{Comparison of the Mean Square Reconstruction Errors (MSRE) per image (followed by the corresponding standard derivations inside the parentheses) of different methods for unseen multi-view faces from the Multi-Pie dataset.}
\newsavebox{\tablebox}
\begin{lrbox}{\tablebox}
  \begin{tabularx}{1\columnwidth}{|c|*{3}{c|}}
   \cline{1-4}
    \diagbox[width=6em]{MSRE}{Methods} & VAE \cite{kingma2013auto} & ABP \cite{han2017alternating} & Ours \\
   \cline{1-4}
    $30^\circ$& $110.99\pm0.11$ & $117.28\pm0.12$ &$\mathbf{89.94\pm0.10}$\\
   \cline{1-4}
    $15^\circ$& $88.98\pm0.09$  & $94.81\pm0.10$   &$\mathbf{70.64\pm0.08}$\\
   \cline{1-4}
     $0^\circ$&$48.78\pm0.05$  & $48.36\pm0.06$    &$\mathbf{46.10\pm0.06}$\\
   \cline{1-4}
    $-15^\circ$&$87.89\pm0.10$ & $94.12\pm0.11$    &$\mathbf{75.11\pm0.09}$\\
   \cline{1-4}
    $-30^\circ$&$107.94\pm0.12$  & $120.58\pm0.13 $ &$\mathbf{92.66\pm0.11}$\\
   \cline{1-4}
    all views&$89.02\pm0.13$ & $94.66\pm0.12 $&$\mathbf{76.52\pm0.10}$\\
   \cline{1-4}
 \end{tabularx}
\end{lrbox}
\scalebox{0.99}{\usebox{\tablebox}}
  \label{table:reconstruction}
\end{table}
\subsection{Experiment 3: Learning on high resolution images}
\begin{figure}[t!]
\begin{center}
\centerline{\includegraphics[width=0.98\columnwidth]{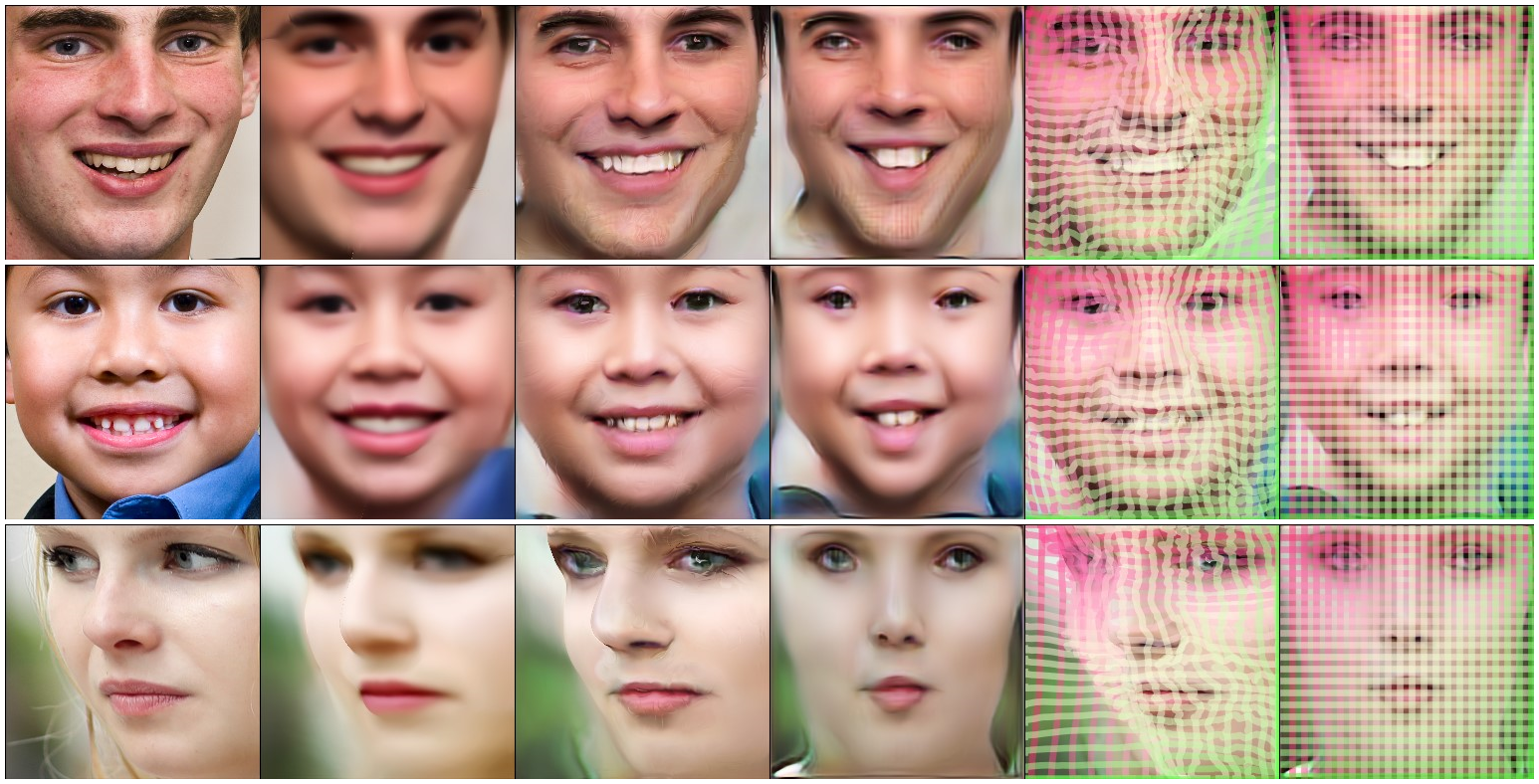}}
\caption{Reconstruction results for high resolution faces. The second column is the reconstruction results by only the deformable generator. The third column is the reconstruction results by recruiting an extra discriminator with adversarial training. The fourth column shows the canonical texture faces outputted by the appearance generator,the last two columns demonstrate the corresponding deformable fields overlaid on the images of the third and fourth columns.}
\label{fig:hirecons}
\end{center}
\vskip -0.1in
\end{figure}

\begin{figure*}[t!]
\begin{center}
\centerline{\includegraphics[width=1.2\columnwidth]{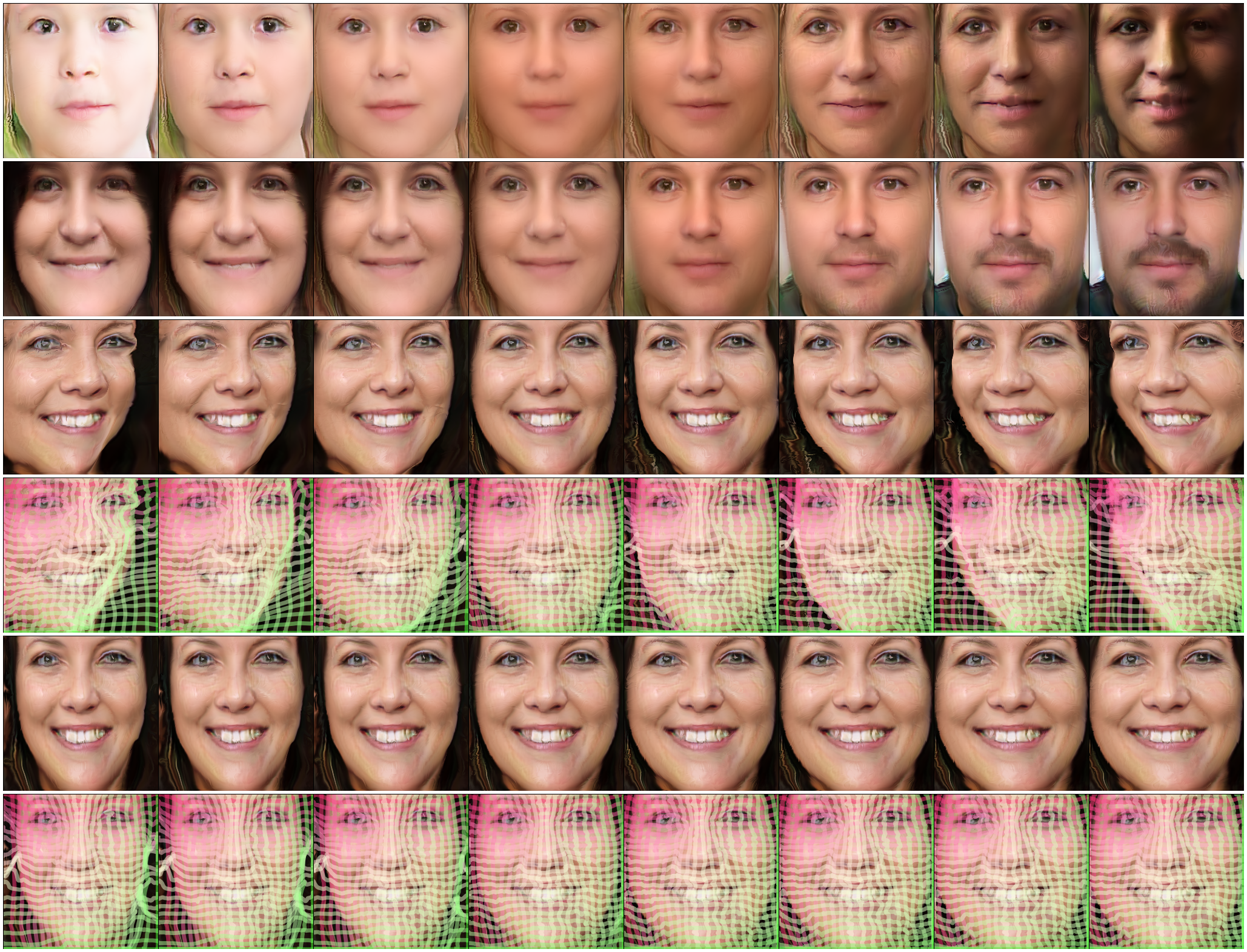}}
\caption{Typical appearance and geometric basis functions for high resolution faces. The appearance basis functions are visualized by the generated images that interpolate the appearance latent factors along the basis functions, while setting geometric latent factor to zeros. The geometric basis functions are visualized by the generated images that interpolate the geometric latent factors along the basis functions, while freezing the appearance latent factor. Refer to section 5.1 for details.}
\label{fig:hibasis}
\end{center}
\vskip -0.1in
\end{figure*}

We conduct experiments on high resolution images. Specifically, we train the deformable generator with 40K faces from FFHQ \cite{karras2020analyzing}, which are cropped to $256 \times 256$ pixels. Since it is well known that the squared Euclidean distance induced from the MLE loss often yields blurry reconstruction and generation results, we recruit adversarial training which includes the deformable generator as an actor and a discriminator which acts as a critic. We extend the MLE loss in Eq. (\ref{eq:MLE}) with the adversarial loss, i.e., the non-saturating loss \cite{goodfellow2014generative} with $R_1$ regularization \cite{mescheder2018training},
\begin{align}
  \nonumber & \min_\theta \max_\phi T(\theta, \phi), \\
    & T(\theta, \phi) = -\lambda_1 L(\theta) + E_{Z^a,Z^g}[\log(1-D(F(Z^a,Z^g;\theta);\phi))]   \label{eq:3kl}\\
\nonumber	&+ E_X[\log D(X;\phi)] + \frac{\lambda_2}{2}E_X[\parallel \nabla D(X;\phi) \parallel^2] ,
\end{align}
where $F(Z^a,Z^g;\theta)$ and  $L(\theta)$ is defined in Eq.(\ref{eq:wholemodel}) and (\ref{eq:MLE}). $D(\cdot;\phi)$ is the discriminator. $\lambda_1=0.001$ and $\lambda_2=10$.
The reconstruction results are illustrated in figure. (\ref{fig:hirecons}). The first column shows the original images. The second column shows the reconstruction results by the MLE loss (Eq. \ref{eq:MLE}) only. The third column shows the reconstruction results by Eq. ( \ref{eq:3kl}) with adversarial loss. The fourth column shows the  canonical texture faces, before warping, learned by the appearance generator. The last two columns demonstrate the corresponding deformable fields overlaid on the images of the third and fourth columns. The adversarial training improves the sharpness of the reconstruction, while the appearance and geometry are still disentangled successfully.
Figure \ref{fig:hibasis} shows the generated basis functions for appearance and geometry. The typical appearance basis functions for illumination and gender changes are shown in the first two rows, while the geometric basis functions for view and shape changes are shown in the rest of four rows with the corresponding deformable grids overlaid. The appearance and geometry are well disentangled. The appearance basis functions only encode front-view information, while the geometric basis functions do not encode any appearance information, that is, the color, illumination and identity are the same across these generated images.

\begin{figure}[t!]
\begin{center}
\centerline{\includegraphics[width=0.9\columnwidth]{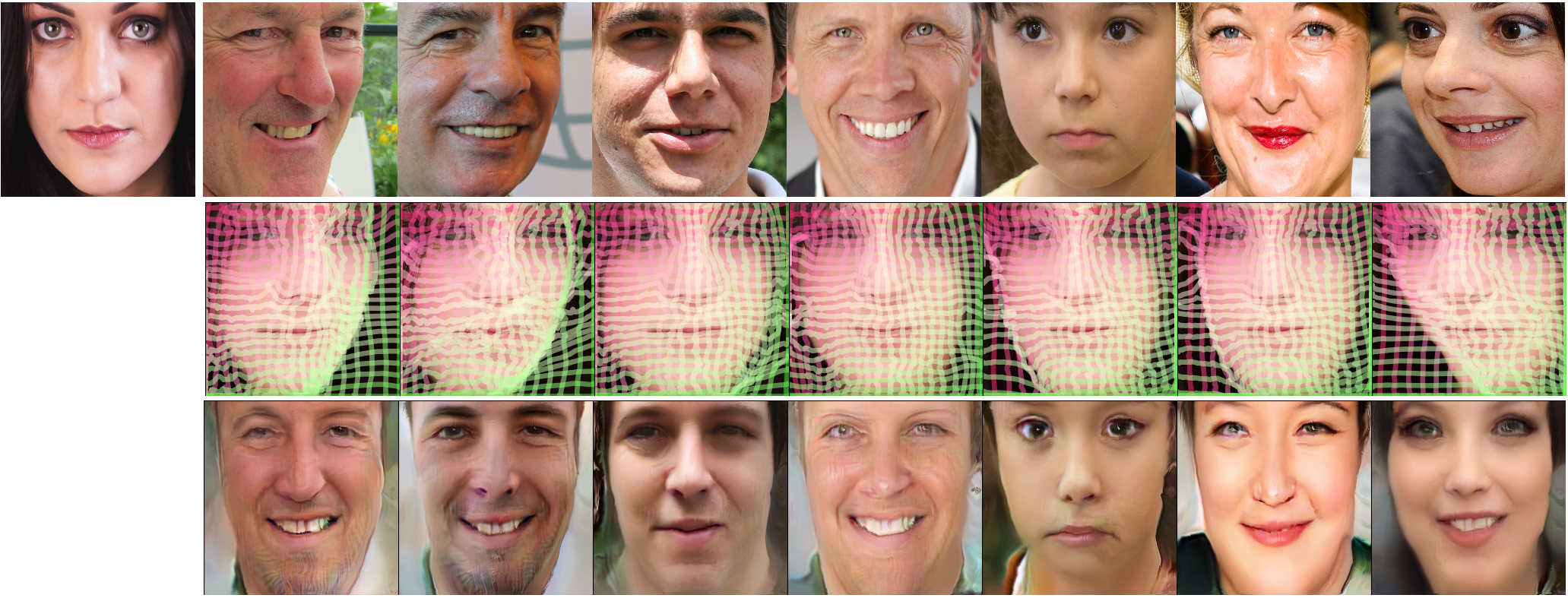}}
\caption{Transferring and recombining geometric and appearance vectors. The first row shows 8 unseen faces from FFHQ. The second row
shows the generated faces by transferring and recombining 2th-8th faces' geometric vectors with first face's appearance vector. The third
row shows the generated faces by recombining the 2th-8th faces' appearance vectors with the first face's geometric vector in the first row.}
\label{fig:hitransfer}
\end{center}
\end{figure}

We further evaluate the ability of disentanglement of the proposed model by transferring and recombining the learned geometric and appearance information from different faces. Specifically, we first feed 8 unseen images from FFHQ into our deformable generator model to infer their appearance vectors $Z^a_1$, $Z^a_2$,$\dots$,$Z^a_8$ and geometric vectors $Z^g_1$, $Z^g_2$ ,$\dots$,$Z^g_8$ using the Langevin dynamics (with 300 steps) in Eq.(\ref{eq:Langevin}). Then, we transfer and recombine the appearance and geometric vectors and use $\{Z^a_1,Z^g_2\}$, $\dots$, $\{Z^a_1,Z^g_8\}$ to generate 7 new face images, as shown in the second row of figure \ref{fig:hitransfer}. We also transfer and recombine the appearance and geometric vectors and use $\{Z^a_2,Z^g_1\}$,$\dots$, $\{Z^a_8,Z^g_1\}$ to generate another 7 new faces, as shown in the third row of figure \ref{fig:hitransfer}. From the 2nd to the 8th column, the images in the second row have the same appearance vector $Z^a$, but the geometric latent vectors $Z^g$ are swapped between each image pair. As shown in the second row of figure \ref{fig:hitransfer}, (1) the geometric information of the original images are swapped in the synthesized images and (2) the inferred $Z^g$ can capture the view information of the unseen images. The images in the third row of figure \ref{fig:hitransfer} have the same geometric vector $Z^g_1$, but the appearance vectors $Z^a$ are swapped between each image pair. From the third row of figure \ref{fig:hitransfer}, we observe that (1) the appearance information are exchanged. (2) The inferred $Z^a$ capture the color, illumination and coarse appearance information but lose more nuanced identity information. Due to the facts that we introduce an extra adversarial loss and the amount of observed images is limited, the model may not be able to represent an arbitrary unseen face accurately.
\subsection{Experiment 4: Learning on non-face dataset}
\begin{figure}[h]
\begin{center}
 \includegraphics[width=\columnwidth]{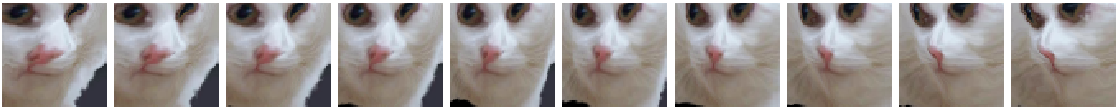}\\
 \includegraphics[width=\columnwidth]{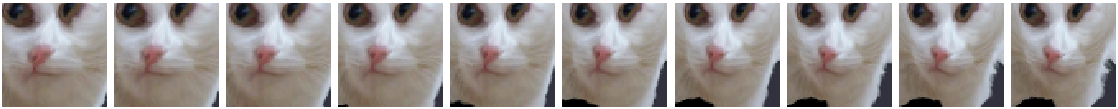}\\
  \includegraphics[width=\columnwidth]{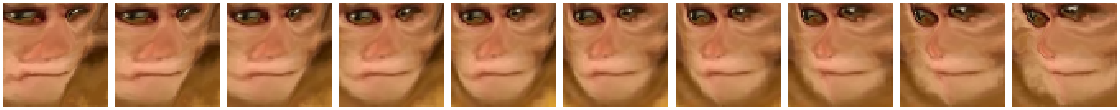}\\
 \includegraphics[width=\columnwidth]{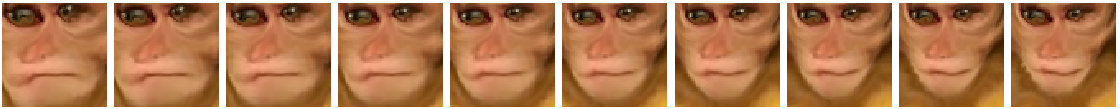}\\
{\footnotesize (a) Transferring the learned geometry from CelebA to animal faces.} \\
 \includegraphics[width=\columnwidth]{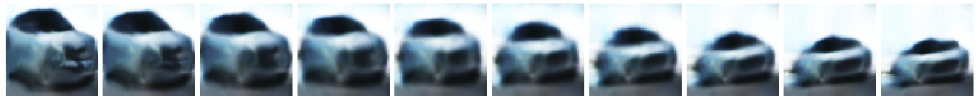}\\
 \includegraphics[width=\columnwidth]{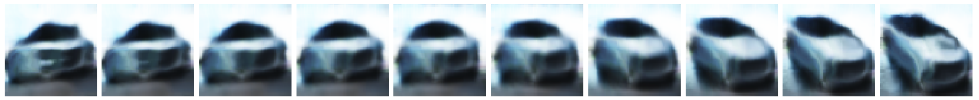}\\
 \includegraphics[width=\columnwidth]{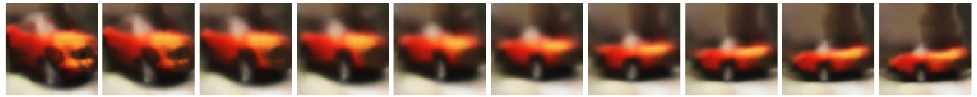}\\
\includegraphics[width=\columnwidth]{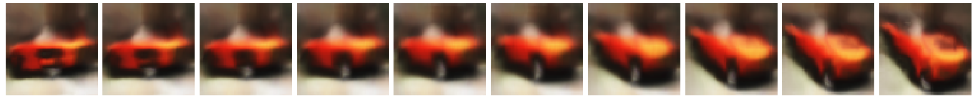}\\
{\footnotesize (b) Geometric interpolations that learned from CIFAR-10. }
\caption{Transferring and learning model from non-face datasets. (a) Geometric interpolation results of cat and monkey faces after applying the rotation and shape warping learned from CelebA. (b) Geometric interpolation results of the model learned from \textit{car} category of CIFAR-10 dataset.}
\label{fig:nonface}
\end{center}
\vskip -0.2in
\end{figure}

We could transfer and learn the model on more general dataset other than face images. For example, the learned geometric information from the CelebA face images can be directly transferred to the faces of animals such as cats and monkeys, as shown in figure \ref{fig:nonface}(a). The cat and  monkey faces rotate from left to right and the shape of the animal faces changes from fat to thin, when the warpings learned from human faces are applied.

We also learn our model on the CIFAR-10 \cite{krizhevsky2009learning} dataset, which includes 50,000 training examples of various object categories. To show the result, we randomly sample and fix $Z^a$ from $\N(0, {\bf I}_{d_a})$. For $Z^g$, we interpolate one dimension from $-\gamma$ to $\gamma$ and fix the other dimensions to $0$. Figure \ref{fig:nonface}(b) shows interpolated examples generated by model learned from the \textit{car} category. The results show that each dimension of $Z^g$ controls a specific geometric transformation, i.e., shape and rotation warping.

\begin{figure*}[t!]
\begin{center}
 \includegraphics[width=1.8\columnwidth]{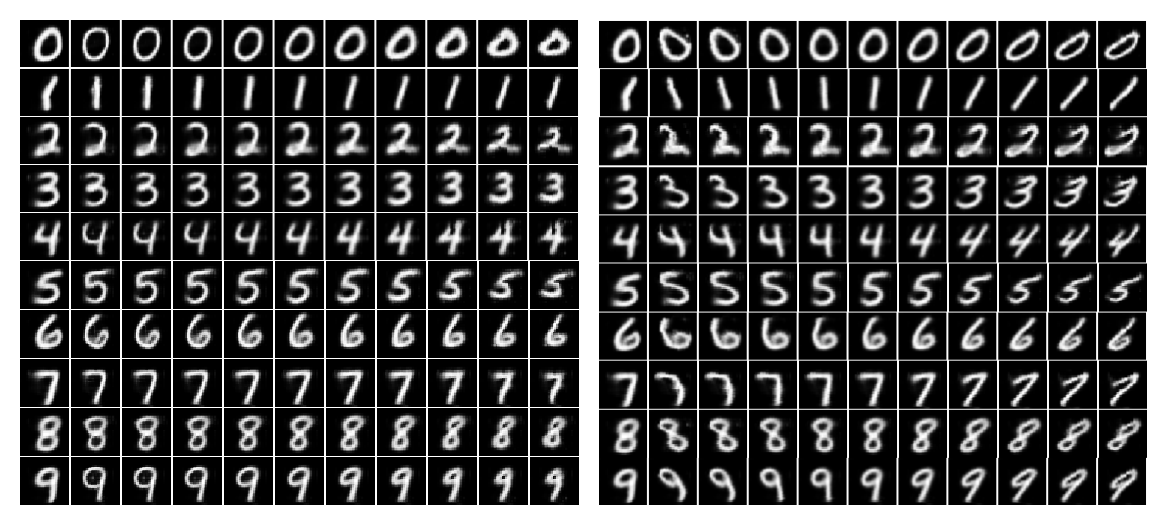}
\caption{Interpolation results by the geometric latent factors of the model learned from the MNIST dataset. Each row demonstrates the results of interpolating one dimension of the geometric latent factor, while keeping $Z^a$ to be one discrete label. In the left subfigure, the first column represent the images generated by the one-hot $Z^a$, and the remain 10 columns show the results by interpolating the shape factor of $Z^g$. As we can observe, from left to right, the shape of the digits change from large to small. Similarly, as we can observe from the right subfigure, the viewing angle of the digits vary from left to right by interpolating the view factor of $Z^g$}
\label{fig:mnists}
\end{center}
\end{figure*}

We next learn our model on the MNIST dataset of 10 classes of digits. In this experiment, $Z^a$ is set to be the prior discrete one-hot label, while $Z^g$ is the continues latent vector. Figure \ref{fig:mnists} demonstrates the interpolated examples. On each row, we set $Z^a$ to be one of the discrete label, while interpolating one dimension of the geometric latent factor $Z^g$ from $[-\gamma,\gamma]$ with a uniform step $\frac{2\gamma}{10}$. In the left subfigure of \ref{fig:mnists}, the first column represent the images generated by the one-hot $Z^a$ (before warping by the deformable fields generated by $Z^g$), and the remain 10 columns show the results by interpolating the shape factor of $Z^g$. As we can observe, from left to right, the shape of the digits change from large to small. Similarly, in the right subfigure of \ref{fig:mnists}, the viewing angle of the digits vary from left to right.
\subsection{Experiment 5: Unsupervised landmark localization}
\begin{figure*}[t!]
\begin{center}
\centerline{\includegraphics[width=1.8\columnwidth]{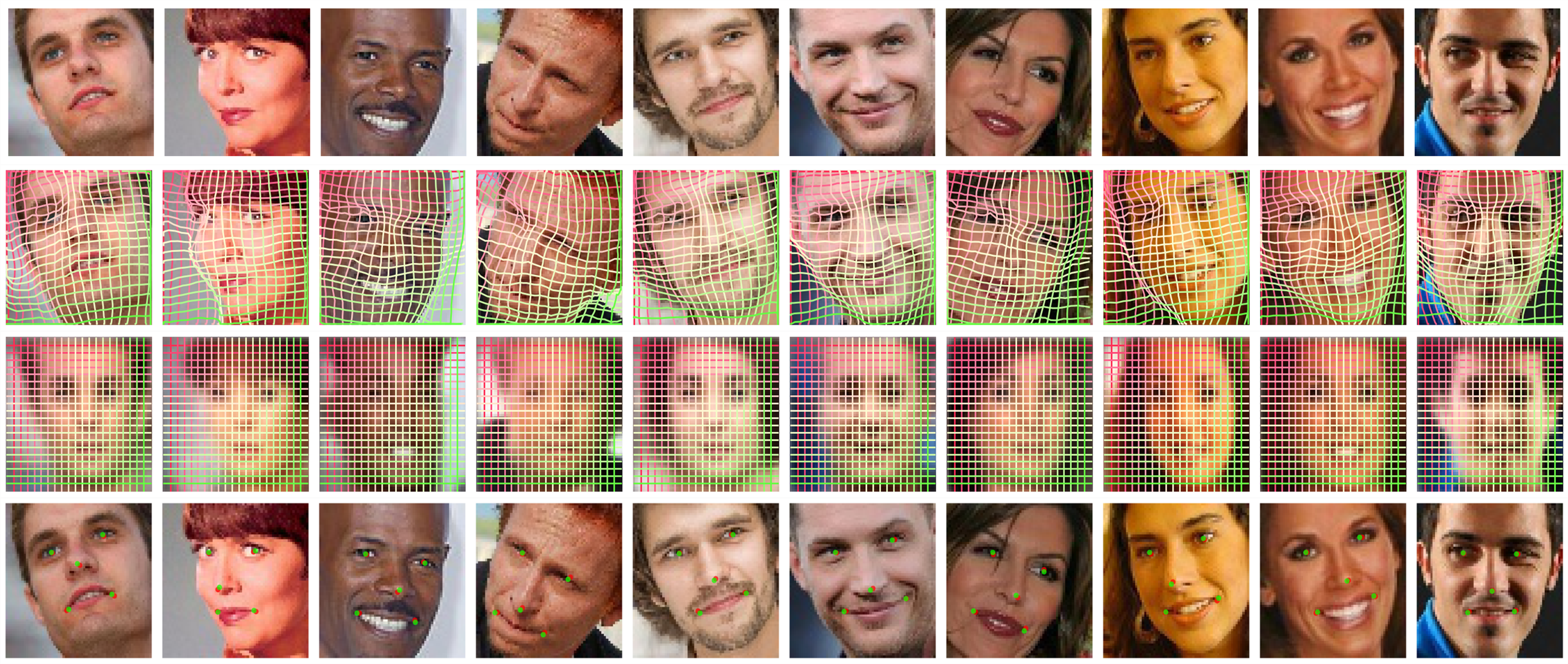}}
\caption{Unsupervised landmark localization. Row 1: the samples of the testing images from the MAFL dataset. Row 2: the deformation grid estimated from warping the the canonical grid with the coordinate displacement (deformation fields) learned from the geometric generator. Row 3: the  canonical grid overlapped on the canonical faces learned from the appearance generator. Row 4:  the semantic landmark locations. The green points denote the ground truth, and the red points denote the predictions.}
\label{fig:unalignldmk}
\end{center}
\end{figure*}
In this subsection, we evaluate the performance of deformable generator model on the task of unsupervised face landmark localization. The experimental results are evaluated on the Multi-Attribute Facial Landmark (MAFL) \cite{zhang2015learning} dataset, which contains landmark locations (eyes, nose, and mouth corners) manually annotated for 19k training and 1k test images.

In this experiment, we train our deformable generator model on the CelebA dataset without any supervision. According to the evaluation protocol of the previous work \cite{thewlis2017unsupervised,shu2018deforming}, employing the provided training annotations in MAFL, we train a landmark regressor post-hoc on the learned deformation fields from the geometric generator. It is worth to mention that the annotation from the MAFL training set is only employed to train the regressor, while our deformable generator model is trained fully unsupervisedly and fixed. The regressor composes of a MLP with 2-layers. The flattened deformation fields (vectors of size $64\times64\times2$), learned from the geometric generator, are fed as input to a hidden layer with 100 neurons, followed the ReLU activation function and an output layer with 10 neurons to predict the 2d spatial coordinates $(x,y)$ for the five landmarks locations (eyes, nose, and mouth corners). The $L_1$ loss is used as the objective function for the landmark regressor.

The first row of figure \ref{fig:unalignldmk} shows the  examples of the testing images. The second row of figure \ref{fig:unalignldmk} demonstrates the estimated deformed grid. The deformation grid is obtained by warping the canonical grid with the deformation fields learned from the geometric generator. The third row of figure \ref{fig:unalignldmk} shows the canonical grid overlaid on the canonical faces learned from the appearance generator. The fourth row of figure \ref{fig:unalignldmk} demonstrates the semantic landmark locations overlapped on the testing images. The green points denote the ground truth, and the red points denote the predictions. As we observe from figure \ref{fig:unalignldmk}, the learned deformation fields could be used for an effective mapping between the landmark locations on the originally unaligned faces and those on the canonical texture faces.

We further quantitatively evaluate the landmark localization by reporting the mean error as a percentage of the inter-ocular distance on the MAFL testing set. We compare our deformable generator model with the other 5 state-of-the-art landmark localization methods \cite{zhang2015learning,thewlis2017unsupervised,shu2018deforming,zhang2014facial}, and report the results on table \ref{table:meulmdk}. As we can observe from table \ref{table:meulmdk}, our method outperforms the other state-of-the-art methods, because our method estimate the deformation field more accurately, even though we never explicitly train the deformable generator to learn correspondence.
\begin{table}[t!]
\centering
		\caption{Comparisons of the mean error of unsupervised landmark prediction on the MAFL test set. Smaller is better.}
		\vspace{3mm}
		\label{table:meulmdk}
 		\begin{lrbox}{\tablebox}
			\begin{tabular}{|c|c|c|c|c|}
				\cline{1-5}
				 TCDCN \cite{zhang2015learning} & Thewlis et al. \cite{thewlis2017unsupervised} &Dense-DAE \cite{shu2018deforming}&MTCNN \cite{zhang2014facial}&Ours\\
				\cline{1-5}
				  7.95  & 5.83 & 5.45 & 5.39 &\textbf{5.18} \\
				\cline{1-5}
			\end{tabular}
 		\end{lrbox}
 		\scalebox{0.85}{\usebox{\tablebox}}
	\end{table}
\subsection{Experiments for Dynamic Deformable Generator}
\begin{figure*}[ht!]
\begin{center}
\includegraphics[width=\columnwidth]{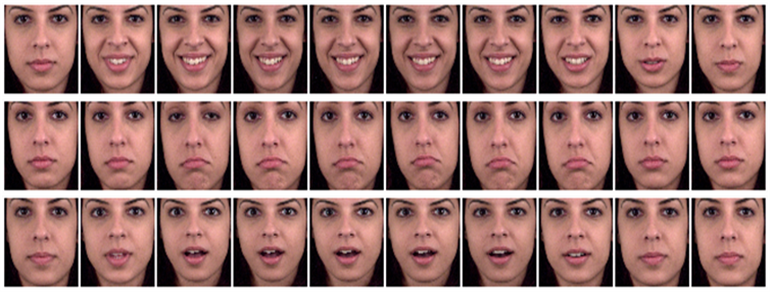}
\includegraphics[width=\columnwidth]{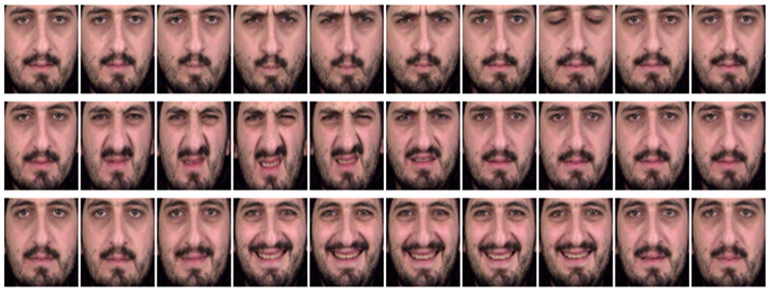}\\
 {\footnotesize (a) Three original video sequences of the same female. $\qquad$ (b) Three original video sequences of the same male} \\~\\
\includegraphics[width=\columnwidth]{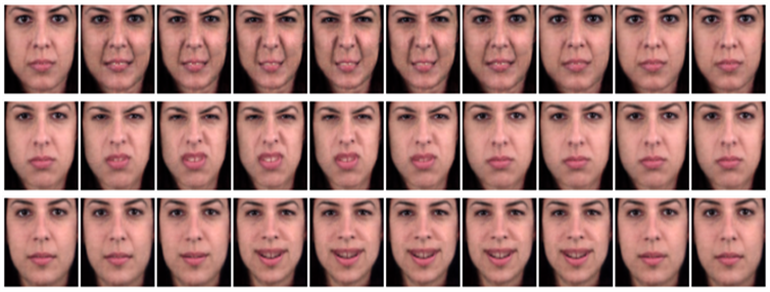}
\includegraphics[width=\columnwidth]{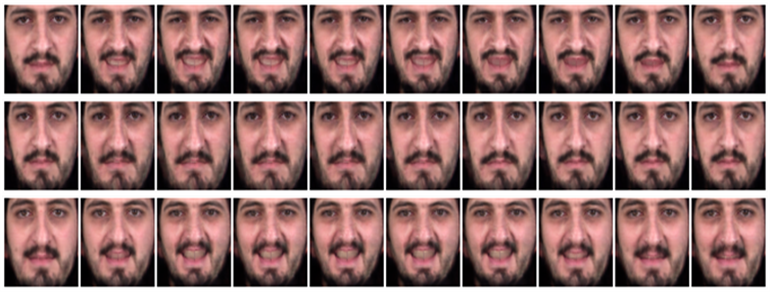}\\
{\footnotesize (c) Recombine the appearance from (a) and geometry from (b).$\qquad \ $  (d) Recombine the appearance from (b) and geometry from (a)}
\caption{Transfer and recombine the appearance and geometric information from different video sequences. (a) Three original video sequences of the same female with expressions of happiness, sadness and surprise; (b) Three original video sequences of the same male with expressions of anger, disgust, and happiness. (c) Transfer and recombine the appearance from female in (a), and the geometry from the male in (b), so that the generated video sequences in (c) inherit the appearance from the female but the shape and expressions (anger, disgust, and happiness) from the male. (d) Transfer and recombine the appearance from the male in (b), and the geometry from the female in (a), so that the generated video sequences in (d) inherit the appearance from the male but the shape and expressions (happiness, sadness and surprise) from the female.}
\label{fig:Mug1}
\end{center}
\end{figure*}

\begin{figure*}[ht!]
\begin{center}
\includegraphics[width=\columnwidth]{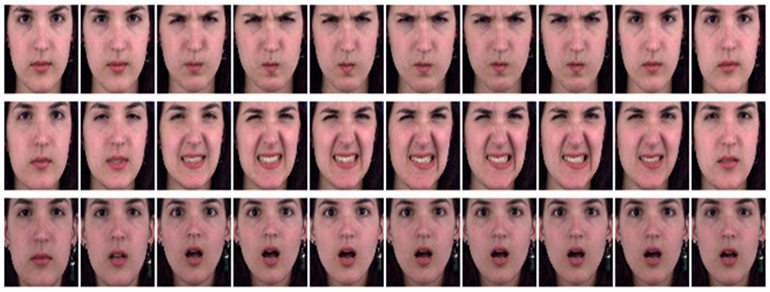}
\includegraphics[width=\columnwidth]{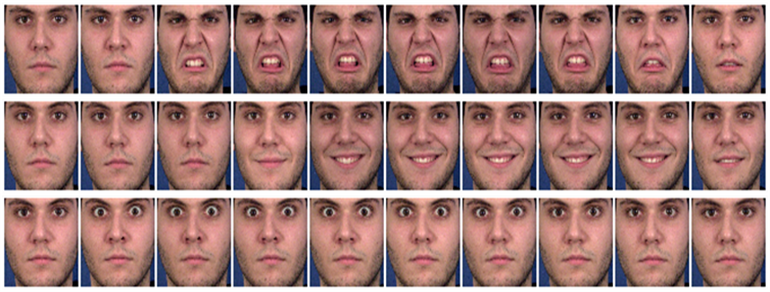}\\
 {\footnotesize (a) Three original video sequences of the same female. $\qquad$ (b) Three original video sequences of the same male}\\~\\
\includegraphics[width=\columnwidth]{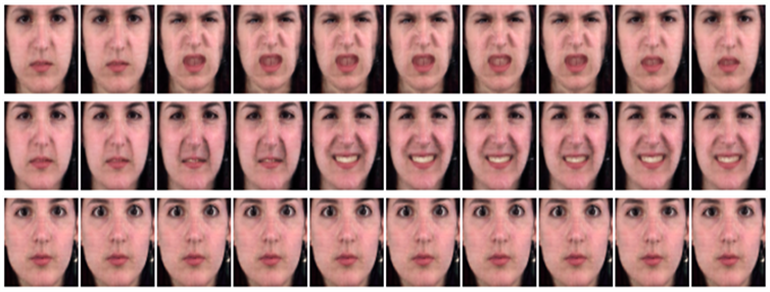}
\includegraphics[width=\columnwidth]{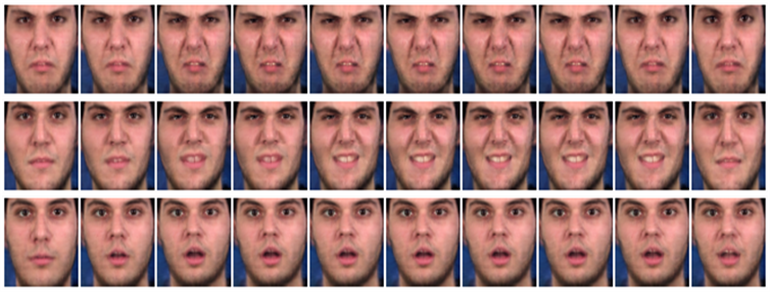}\\
{\footnotesize (c) Recombine the appearance from (a) and geometry from (b).$\qquad \ $  (d) Recombine the appearance from (b) and geometry from (a)}
\caption{Transfer and recombine the appearance and geometric information from different video sequences. (a) Three original video sequences of the same female with expressions of anger,disgust and surprise; (b) Three original video sequences of the same male with expressions of disgust, happiness, and fear. (c) Transfer and recombine the appearance from female in (a), and the geometry from the male in (b), so that the generated video sequences in (c) inherit the appearance from the female but the shape and expressions (disgust, happiness, and fear) from the male. (d) Transfer and recombine the appearance from the male in (b), and the geometry from the female in (a), so that the generated video sequences in (d) inherit the appearance from the male but the shape and expressions (anger,disgust and surprise) from the female.}
\label{fig:Mug2}
\end{center}
\end{figure*}

To study the performance of the proposed dynamic deformable generator in disentangling the appearance and geometric information from the video sequences, we experiment on the MUG \cite{5617662} facial expression video dataset.  The dataset consisted of 86 subjects. The video sequences of each subject represent one of the six facial expressions: happiness, sadness, anger, fear, disgust, and surprise.
We crop the face regions by the OpenFace and scaled to $64\times 64$ pixels $\times 60$ frames. Some example video sequences of MUG are shown in the first 3 rows of figure \ref{fig:Mug1} and figure \ref{fig:Mug2}.

To evaluate the performance of disentanglement on the video sequences, we first consider the experiment on transferring and recombining the
appearance and geometric information from different video sequences. Specifically, consider two video sequences from different persons. We first learn and infer the sequences of appearance and geometric latent factors from the two videos as $\{Z^a_{1,t},Z^g_{1,t}, t=0,\dots, T\}$ and $\{Z^a_{2,t},Z^g_{2,t}, t=0,\dots, T\}$. Then we transfer and recombine them to generate new video sequences.  More specifically, we expect that recombining the sequence of appearance latent factors from the the first person and the sequence of geometric latent factors from the second person, $\{Z^a_{1,t},Z^g_{2,t}, t=0,\dots, T\}$, will generate a new video that inherits the appearance information from the first person, while inherits the dynamic geometric information (including both the face shape and dynamic face expression) from the second person. Similarly, we also expect that  recombining the sequence of appearance latent factors from the the second person and the sequence of geometric latent factors from the first person, $\{Z^a_{2,t},Z^g_{1,t}, t=0,\dots, T\}$, will generate another new video that inherits the appearance information from the second person, while inherits the dynamic geometric information from the first person.

Figure \ref{fig:Mug1} demonstrates the experimental results on transferring and recombining the appearance and geometric information from different video sequences.  Figures \ref{fig:Mug1} (a) and (b) show 6 different expressional video sequences of a female and a male. Each row of figure \ref{fig:Mug1} (c) demonstrates a generated video sequence by recombining the sequence of appearance latent factors from the same row of the figure \ref{fig:Mug1} (a) and the sequence of geometric latent factors of the same row from the figure \ref{fig:Mug1} (b).  Figure \ref{fig:Mug1} (d) demonstrates the generated videos sequences by recombining the sequence of appearance latent factors from the male and the sequence of geometric latent factors from the female.

As we observe from figure \ref{fig:Mug1} (c), the appearance information, such as the face color, texture, and the five senses, are the same as that of the female, while the geometric information, such as the face shape and dynamic face expression, are similar with the that of the female. In figure \ref{fig:Mug1} (d), the appearance information are the same as that of the male, while the face shape and dynamic face expression are similar with the that of the female. More specifically, comparing figure \ref{fig:Mug1} (d) and figure \ref{fig:Mug1} (b), we can observe that the face shape in figure \ref{fig:Mug1} (d) is obviously thinner than that of the original male's face shape as shown in Subfigure \ref{fig:Mug1} (b), which inherits the face shape of the female from figure \ref{fig:Mug1} (a). Comparing figure \ref{fig:Mug1} (c) and figure \ref{fig:Mug1} (a), we can observe that although the face shape and dynamic face expression are changed, the face color, texture, and the detailed appearance features, such as the eyebrows are kept the same as the original female's eyebrows.

Figure \ref{fig:Mug2} also demonstrates the experimental results on transferring and recombining the appearance and geometric information from video sequences of another two persons. From figure \ref{fig:Mug2}, we observe similar phenomena as that from figure \ref{fig:Mug1}. The above experimental results verify that the appearance and geometric information from the video sequences are well disentangled by the proposed dynamic deformable generator model.

\begin{figure}[t]
\vskip -0.02in
\begin{center}
 \includegraphics[width=\columnwidth]{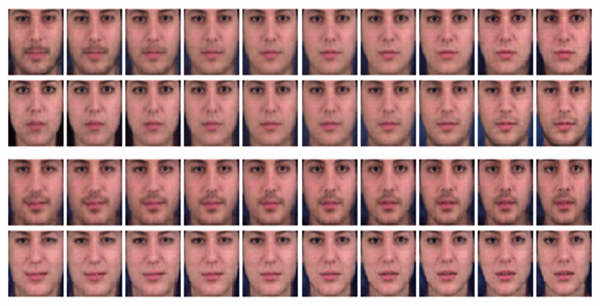}
\caption{The first two rows show the typical appearance basis functions for the first frame, visualized by the generated images from interpolating the appearance latent factors along the basis functions. Each dimension of the appearance latent factors encodes appearance information such as gender and background color. The last two rows demonstrate the major geometric basis function, visualized by applying it over the appearance basis functions in the first two rows. The major geometric basis function captures the shape information. From left to the right, the shape of the faces at the last two rows change from fat to thin.}
\label{fig:appgeov}
\end{center}
\end{figure}

Next, we further study disentangling of the learned geometric information from the video into the global shape and dynamic expression. Thus, we can edit the face shape of a person, while keeping his dynamic face expression from the original video. Recall in Eq. (\ref{eq:latentsum}) in Section 4,  the sequence of geometric latent factors $Z^g_t, \{t=0,\dots,T\}$ can be represented by the summation, in the latent space, of the geometric latent factor of the first frame $Z^g_0$ and the consequent hidden state vectors $s^g_t, \{t=1,\dots,T\}$. Thus, the geometric latent factor of the first frame can be employed to represent the global shape, while the consequent hidden state vectors can be utilized to represent the dynamic expression. We plot the learned appearance basis functions for the first frame in the first two rows of figure (\ref{fig:appgeov}), according to the method we did in Section 5.1. By observing the interpolating results of the appearance latent factors along the basis functions among $[-\gamma,\gamma]$, we find the major appearance basis functions capture the identity and color changing of the person, such as identity varying from male to female (the first row), and the background color varying from black to blue (the second row). Since the MUG dataset only contains face expression videos from the front view, the major geometric basis function captures the shape information. The third and the fourth rows of figure (\ref{fig:appgeov}) demonstrate the major shape basis function by applying it over the appearance basis functions in the first two rows when keeping the interpolation value equals to $\frac{-4\gamma}{5}$. As we can observe from the interpolation results of the geometric latent factors $Z^g_0$ along the basis functions among $[-\gamma,\gamma]$, the face shape varies from fat to thin, while the identity information from the appearance is kept invariant.

\begin{figure}[t]
\begin{center}
 \includegraphics[width=0.96\columnwidth]{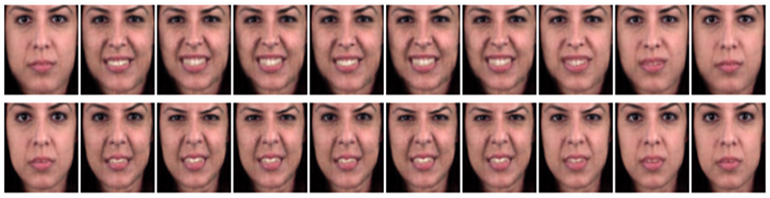}\\
 \includegraphics[width=0.96\columnwidth]{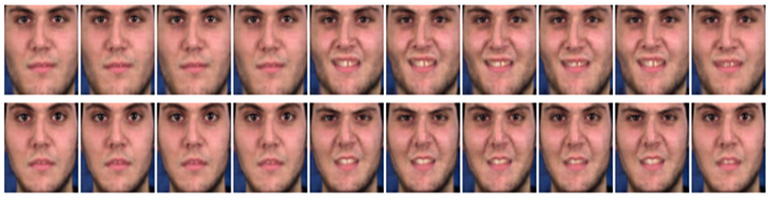}\\
\caption{Editing the shape of the whole video sequence of facial expression. The face shapes of the video sequences in the second and the fourth row are thinner than that of the first and the third row, while the facial expressions, as well as the appearance information, of the video sequences in the second and the fourth row are kept invariant as that of the first and the third row.}
\label{fig:appshapev}
\end{center}
\vskip -0.3in
\end{figure}

To edit the global shape information of the video sequence, we can recombine the geometric latent factors $Z^g_0$ with different coefficients corresponding to the shape basis function, and the consequent hidden state vectors $s^g_t, \{t=1,\dots,T\}$ to generate a new sequence of geometric latent factors $Z^g_t, \{t=0,\dots,T\}$. Figure (\ref{fig:appshapev}) shows the results of editing the shape of two face expression videos. As we can observe from figure (\ref{fig:appshapev}), the face shapes of the second row and the fourth row are thinner than that of the first and the third row, while the facial expressions, as well as the appearance information, of the second row and the fourth row are the same as that of the first and the third row. These results verify that the proposed dynamic deformable generator model can not only disentangle the appearance and geometric information of the video sequence, but also further disentangle the global shape and dynamical expression among the geometric information.

\subsection*{Dynamic Deformable fields for facial expression analysis and recognition}
\begin{figure}[t]
\begin{center}
 \includegraphics[width=0.96\columnwidth]{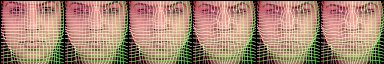}\\
 \includegraphics[width=0.96\columnwidth]{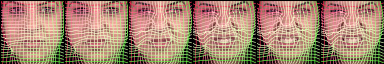}\\
  \includegraphics[width=0.96\columnwidth]{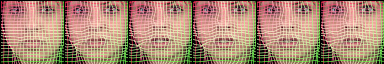}\\
   \includegraphics[width=0.96\columnwidth]{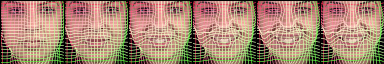}\\
   \includegraphics[width=0.96\columnwidth]{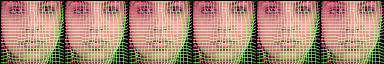}\\
   \includegraphics[width=0.96\columnwidth]{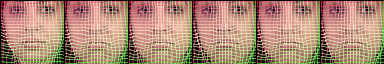}\\
    \includegraphics[width=0.96\columnwidth]{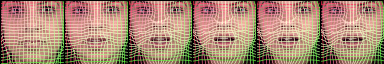}
\caption{Dynamic deformed grids for 7 different facial expressions over the corresponding facial sequences. From the top row to the bottom row, the facial expressions are anger, disgust, fear, happiness, neutral, sadness. and surprise. As can be observed from the regions around the eyes, brows, nose, and mouth, the learned dynamic deformation fields (demonstrated as  deformed grids) reflect the intrinsic feature of different kinds of facial expressions. See text for detail. Best viewed with magnification.}
\label{fig:dfvexpression}
\end{center}
\vskip -0.3in
\end{figure}

We next demonstrate that the learned dynamic deformable fields can be used for facial expression analysis and recognition. To visualize the dynamic deformable fields, we plot the dynamic deformed grids over the corresponding facial sequences. As we can observe from figure \ref{fig:dfvexpression}, for different expressions, the dynamic deformed grids are different. The first row of figure \ref{fig:dfvexpression} demonstrate the anger. As observed from the deformable lines around the eyes, brows, and mouth, the implicit movements are inward lowering the brows and mouth compaction. The second row demonstrate the disgust.  As observed from the deformable lines around the nose, mouth and brows, the implicit movements are upward nose motion, mouth expanded and opened, and lowering of brows. The third row show the fear.  As observed from the deformable lines around the mouth and brows, the implicit movements are slight expansion and raising of mouth and raising inner parts of brows. The fourth row shows the happiness.  As observed from the deformed grids around the mouth,  the implicit movement is mouth opening with its expansion. 
The sixth row shows the sadness. As observed from the deformable lines around the mouth, the implicit movement is lowing mouth corners and raising mid mouth. The last row demonstrate the surprise. As observed from the grid lines around the brows and the lip, the implicit movement is raising brows and raising the upper lip.

Since the facial expression is connected with the dynamic geometric information and unrelated with the appearance information, such as color, illumination, and identity, we can employ both the learned dynamic geometric latent factors $\{Z^g_t, t=0,\cdots,T \}$ and the dynamic deformation fields $\{F_g(Z_t^g;\theta_g),t=0,\cdots,T \}$ (defined in Eq. \ref{eq:dynamodel}) as the feature to recognize the facial expressions. Specifically, the learned dynamic geometric latent factors (dglf) or the dynamic deformable fields (ddf) are fed into the long short-term memory (LSTM) to model the temporal change. More specifically, we utilize an one-layer LSTM, and the output of LSTM is connected with a fully connected layer with 30 hidden neurons and a soft-max layer with cross-entropy loss. To evaluate the performance of our method, we compare our method with MoCoGAN \cite{tulyakov2018mocogan}, whose motion latent factors are feed to the same LSTM and the logistic classifier. We also compare our method with 5 state-of-art facial emotion recognition algorithms: DAGSVM-GT \cite{sen2019facial}, UPSM \cite{weber2018unsupervised}, HiNet \cite{verma2019hinet}, Multi-stream CNN \cite{aghamaleki2019multi}, and RADAP \cite{mandal2019regional}. According to the experimental setup in RADAP, UPSM and HiNet, in our experiments, the 10-fold person independent cross-validation scheme is employed.

In table \ref{table:emotionrec}, we measure the performance of facial emotion recognition in terms of average recognition accuracy. Our method using dynamic geometric latent factors (dglf) performs better than the MoCoGAN's motion latent factors as the feature representation. Moreover, the learned dynamic deformable fields (ddf) achieves superior recognition accuracy compared with the other algorithms. The reasons are as follows: (1) The disentangled dynamic geometric information is more effective for facial expression recognition, since the appearance information, such as color, illumination, and identity, are unrelated with the expression or the emotion and sometimes may result in negative effects. (2) Our model disentangles the appearance and geometric information by two generators connected by the warping function. The geometric warping only modifies the positions of pixels in an image without changing the color or illumination. Therefore, the extracted dense deformable fields is more effective and pure than the motion latent factors learned by the MoCoGAN, which disentangles the content and motion latent factor from one concatenated latent vector. (3) The learned dense deformable fields contain more important information than the methods based on sparse landmarks such as UPSM \cite{weber2018unsupervised} and DAGSVM-GT \cite{sen2019facial}.

\begin{table*}[t!]
\centering
		\caption{Comparisons of the mean recognition accuracy of  seven different methods on the MUG dataset. Dglf and Ddf denote our method employ the learned dynamic geometric latent factors and the dynamic deformable fields as the feature representation, respectively.}
		\vspace{3mm}
		\label{table:emotionrec}
			\begin{tabular}{|c|c|c|c|c|c|c|c|}
				\cline{1-8}
				 DAGSVM-GT\cite{sen2019facial}& UPSM \cite{weber2018unsupervised} &HiNet \cite{verma2019hinet}&Multi-stream CNN \cite{aghamaleki2019multi}& RADAP \cite{mandal2019regional} &MoCoGAN \cite{tulyakov2018mocogan}& Our's Dglf & Our's Ddf  \\
				\cline{1-8}
				  82.3  &  87.7 & 87.2 & 85.4 & 85.6 & 83.1 & 85.5 &\textbf{92.0} \\
				\cline{1-8}
			\end{tabular}
	\end{table*}
\subsection{Balancing explaining-away competition and Network Structures}
The proposed deformable generator model utilizes two generator networks to disentangle the appearance and geometric information from an image. Since the geometric generator only produces displacement for each pixel without modifying the pixel's value, the color and illumination information and the geometric information are naturally disentangled by the proposed model's specific structure.

In order to properly disentangle the identity (or category) and the view (or geometry) information, the learning capacity between the appearance generator and geometric generator should be balanced. The appearance generator and the geometric generator cooperate with each other to generate the images. Meanwhile, they also compete against each other to explain away the training images. If the learning of the appearance generator outpaces that of the geometric generator, the appearance generator will encode most of the knowledge (including the view and shape information), while the  geometric generator will only learn minor warping operations. On the other hand, if the geometric generator learns much faster than the appearance generator, the geometric generator will encode most of the knowledge (including the identity or category information), which should be encoded by the appearance network.

To control the tradeoff between the two generators, we propose two schemes. In the first  scheme, we introduce a balance parameter $\alpha$, which is defined as the ratio of the number of filters within each layer between the appearance and geometric generators. The balance parameter $\alpha$ should not be too large or too small. In the experiments of deformable generator for images, we adopt this scheme, and set $\alpha$ to 0.625 in our experiments. The appearance and geometric generator's structures are shown in Table \ref{table:networkimg}.
\begin{table}
	\centering
	\caption{Network architectures of deformable generators for images. The kernel size of the deconvolution layer is listed in $ [\cdot ]$ and the number of output channel is listed in $(\cdot)$. The stride of all the deconvolution layer is 2. The balance parameter $\alpha$ is set to 0.625}
	\vspace{2mm}
	\label{table:networkimg}
	\scriptsize
	\begin{tabular}{c|c||c }
		\hline
		Layer  & Appearance Generator Network & Geometric Generator Network \\ \hline
		1 &  $Z^a\sim \mathcal{N}(0, I_{64})$ & $Z^g\sim \mathcal{N}(0, I_{64})$ \\ \hline
		2 & FC, $(4\times 4\times 128 \times \alpha)$ & FC, $(4\times 4\times 128)$\\ \hline
	    3 & Deconv+ReLU, $[3\times 3],(64 \times \alpha)$  &  Deconv+ReLU, $[3\times 3],(64)$ \\ \hline
        4 & Deconv+ReLU, $[3\times 3],(32 \times \alpha)$  &  Deconv+ReLU, $[3\times 3],(32)$ \\ \hline
        5 & Deconv+ReLU, $[5\times 5],(16 \times \alpha)$  &  Deconv+ReLU, $[5\times 5],(16)$ \\ \hline
		6 & Deconv+Tanh, $[5\times 5],(3)$ &  Deconv+Tanh, $[5\times 5],(2)$ \\ \hline
				
	\end{tabular}
\end{table}

In the second scheme,  we explicitly scale the weights between the two generators with a balance parameter $\beta$ at runtime to control the tradeoff. Specifically, we set $\hat{w}^a_i =  w^a_i / c $ and $\hat{w}^g_i =  w^g_i / c \times \beta$, where $w^a_i$ and $w^g_i$ are the weights for the appearance and geometric generators, $c$ is the per-layer normalization constant from He's initializer \cite{he2015delving}. This dynamical updating scheme for scaling the two set of weights can control the tradeoff between the two generators more flexible, because, in scheme 1, the number of filter is an integer, which results in the limited choice of the balance parameter $\alpha$, while, in scheme 2, the values of the weights are decimals which results in the infinite number of the choice for the balance parameter $\beta$. In the experiments of the dynamic deformable generator for the video sequences, we adopt the second scheme, and set $\beta$ to 0.3 in our experiments. The corresponding appearance and geometric generator's structures are shown in Table \ref{table:networkvideo}.
\begin{table}[h!]
	\centering
	\caption{Network architectures of dynamic deformable generators for video sequences. The kernel size of the deconvolution layer is listed in $ [\cdot ]$ and the number of output channel is listed in $(\cdot)$. The stride of the appearance generator network's first 4 deconvolution layer is 2, for the last deconvolution layer is 1 . The stride of the geometric generator network's first 2 deconvolution layer is 4, for the last deconvolution layer is 1. The balance parameter $\beta$ is set to 0.3}
	\vspace{2mm}
	\label{table:networkvideo}
	\scriptsize
	\begin{tabular}{c|c||c }
		\hline
		Layer  & Appearance Generator Network & Geometric Generator Network \\ \hline
        1  &  $Z^a_0 \sim \mathcal{N}(0, I_{60}), s^a_0=0$ & $Z^g_0 \sim \mathcal{N}(0, I_{30}), s^g_0=0$\\ \hline
		2 & FC+ReLU, $(4\times 4\times 512)$ & FC+ReLU, $(4\times 4\times 256)$\\ \hline
	    3 & Deconv+ReLU, $[3\times 3],(512)$  &  Deconv+ReLU, $[4\times 4],(128)$ \\ \hline
        4 & Deconv+ReLU, $[3\times 3],(256)$  &  Deconv+ReLU, $[4\times 4],(64)$ \\ \hline
        5 & Deconv+ReLU, $[3\times 3],(128)$  &    Deconv+Tanh, $[5\times 5],(2)$ \\ \hline
		6 & Deconv+ReLU, $[3\times 3],(64)$ &   \\ \hline
        7 & Deconv+Tanh, $[5\times 5],(3)$ &   \\ \hline \hline
		Layer  & Appearance Transition Network & Geometric Transition Network \\ \hline
     	1 & $\xi_t^a \sim \mathcal{N}(0, I_{3}), s^a_0=0$  &  $\xi_t^g \sim \mathcal{N}(0, I_{10}), s^g_0=0$ \\ \hline	
        2 & FC+ReLU, (20) & FC+ReLU, (20) \\ \hline	
        3 & FC+ReLU, (20) & FC+ReLU, (20) \\ \hline	
        4 & FC+tanh, (60) & FC+tanh, (30) \\ \hline	
	\end{tabular}
\end{table}
\section{Conclusion and future work}
In this study, we propose a deformable generator model which aims to disentangle the appearance and geometric information of an image into two independent latent vectors $Z^a$ and $Z^g$. We also introduce a dynamic deformable generator model for the spatial-temporal process which disentangle the appearance and geometric information of a video sequence into two groups of independent latent vectors $Z^a_t$ and $Z^g_t$. The learned geometric generator can be transferred to other datasets that share similar structure regularity, or can be used for the purpose of data augmentation to produce more variations in the training data for better generalization. 

The proposed method models the 2D displacement field of the image grid explicitly. Currently the method is limited by local displacement field or warping. In our future work, we shall study more global displacements of image grid caused by the changes of 3D poses of the objects as well as camera viewpoint. Such 3D changes may act on the image intensities through the 2D displacement field in our model, instead of acting on the image intensities directly, so that the mapping from the 3D changes to the 2D displacement field can be much simpler than the mapping from the 3D changes to the image intensities directly. Thus it is possible to combine the proposed method with the recent GAN methods \cite{karras2019style,karras2020analyzing,nguyen2019hologan} to make the model more explicit and simpler.


In our work, we mainly focus on disentangling the overall appearance and geometric information into two (groups of) independent vectors. We may further introduce other regularization terms, such as the mutual information or the total correlation, into the existing training objective to enforce and encourage the latent factors within the appearance group or geometry group to be more interpretable and independent.


In the current work, we focus on extracting the appearance and geometric knowledge from the image or video data. For more general situation, we may further extract the topology information, and thus we will disentangle the appearance, geometry and topology information all together. The topology information may take care of specific structure information. Specifically, different categories of images usually take on different structures, so we can employ the one-hot discrete latent variables to model this kind of topology (or structure) information.  From the perspective of unsupervised clustering, by disentangling the topology information, we can cluster the data with similar topology information. 
 \section*{Acknowledgment}
Part of the work was done while the first author was visiting UCLA as a visiting scholar. The work of the first author was supported by the Natural Science Foundation of China No. 61703119, the CAAI-Huawei MindSpore Open Fund No. CAAIXSJLJJ-2020-033A, the Natural Science Fund of Heilongjiang Province of China No. QC2017070, and Fundamental Research Funds for the Central Universities No. 3072020CF0403. The work of the other authors was supported by NSF DMS-2015577, DARPA XAI N66001-17-2-4029, ARO W911NF1810296, and ONR MURI N00014-16-1-2007.


{\small
	\bibliographystyle{IEEEtran}
	\bibliography{deformable-PAMI}
}
\end{document}